\documentclass{article}
\usepackage{microtype}
\usepackage{lineno}
\usepackage{graphicx}
\usepackage{subcaption}
\usepackage{booktabs} 
\usepackage{hyperref}
\usepackage{multicol, multirow, graphicx}
\usepackage{array}
\usepackage{tabularx}
\usepackage{wrapfig}

\usepackage[accepted]{icml2025}

\usepackage{amsmath}
\usepackage{amssymb}
\usepackage{mathtools}
\usepackage{amsthm}
\usepackage{booktabs}

\usepackage[capitalize,noabbrev]{cleveref}
\usepackage{tikz} 
\usepackage{colortbl, xcolor}
\usepackage{kotex}
\usepackage{bbding}
\usepackage{pifont}
\usepackage{caption}
\theoremstyle{plain}
\newtheorem{theorem}{Theorem}[section]

\newtheorem{lemma}[theorem]{Lemma}

\theoremstyle{definition}

\theoremstyle{remark}

\usepackage[textsize=tiny]{todonotes}

\newcommand{\cmark}{\ding{51}}
\newcommand{\xmark}{\ding{55}}
\definecolor{tabfirst}{rgb}{1, 0.7, 0.7}
\definecolor{tabsecond}{rgb}{1, 0.85, 0.7}
\definecolor{tabthird}{rgb}{1, 1, 0.7}

\icmltitlerunning{Skrr: Skip and Re-use Text Encoder Layers for Memory Efficient Text-to-Image Generation}

\begin{document}

\twocolumn[
\icmltitle{Skrr: Skip and Re-use Text Encoder Layers for \\ Memory Efficient Text-to-Image Generation}



\icmlsetsymbol{equal}{*}

\begin{icmlauthorlist}
\icmlauthor{Hoigi Seo}{equal,ece}
\icmlauthor{Wongi Jeong}{equal,ece}
\icmlauthor{Jae-sun Seo}{cornell}
\icmlauthor{Se Young Chun}{ece,inmc}
\end{icmlauthorlist}

\icmlaffiliation{ece}{Dept. of Electrical and Computer Engineering, Seoul National University, Republic of Korea}
\icmlaffiliation{cornell}{School of Electrical and Computer Engineering, Cornell Tech, USA}
\icmlaffiliation{inmc}{INMC \& IPAI, Seoul National University, Republic of Korea}

\icmlcorrespondingauthor{Se Young Chun}{sychun@snu.ac.kr}

\icmlkeywords{Generative model, Efficiency, Pruning}

\vskip 0.3in
]



\printAffiliationsAndNotice{\icmlEqualContribution} 

\begin{abstract}
Large-scale text encoders in text-to-image (T2I) diffusion models have demonstrated exceptional performance in generating high-quality images from textual prompts. Unlike denoising modules that rely on multiple iterative steps, text encoders require only a single forward pass to produce text embeddings. However, despite their minimal contribution to total inference time and floating-point operations (FLOPs), text encoders demand significantly higher memory usage, up to eight times more than denoising modules. To address this inefficiency, we propose Skip and Re-use layers (Skrr), a simple yet effective pruning strategy specifically designed for text encoders in T2I diffusion models. Skrr exploits the inherent redundancy in transformer blocks by selectively skipping or reusing certain layers in a manner tailored for T2I tasks, thereby reducing memory consumption without compromising performance. Extensive experiments demonstrate that Skrr maintains image quality comparable to the original model even under high sparsity levels, outperforming existing blockwise pruning methods. Furthermore, Skrr achieves state-of-the-art memory efficiency while preserving performance across multiple evaluation metrics, including the FID, CLIP, DreamSim, and GenEval scores.
\end{abstract}

\section{Introduction}
\label{intro}
Diffusion generative models excel at tasks such as text-to-image (T2I) synthesis~\cite{rombach2022high, podell2023sdxl, esser2024scaling, chen2025pixart}, editing~\cite{brooks2023instructpix2pix, cao2023masactrl, kawar2023imagic}, video generation~\cite{liu2402sora, polyak2024movie}, and 3D creation~\cite{poole2022dreamfusion, cao2023masactrl, seo2023ditto, wang2024prolificdreamer}. With modern architecture and large-scale text encoders, they produce high-quality images that closely match text prompts. Despite these successes, they require significant computational resources, especially memory, making deployment and scalability challenging.

\begin{figure}[!t]
  \centering
  \begin{subfigure}{0.495\linewidth}
    \includegraphics[width=1\linewidth]{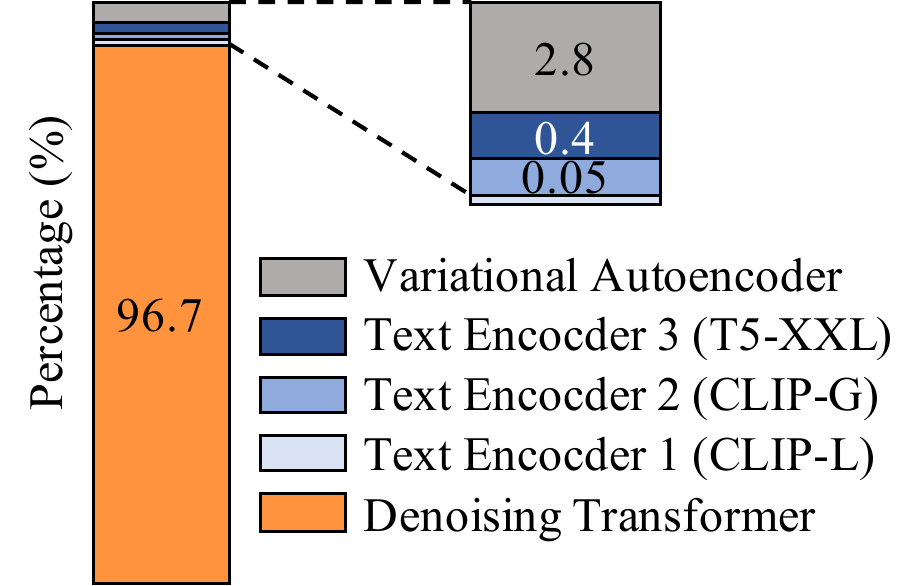}
    \caption{FLOPs ratio.}
    \label{fig:parameter_ratio}
  \end{subfigure}
  \hfill
  \begin{subfigure}{0.495\linewidth}
    \includegraphics[width=1\linewidth]{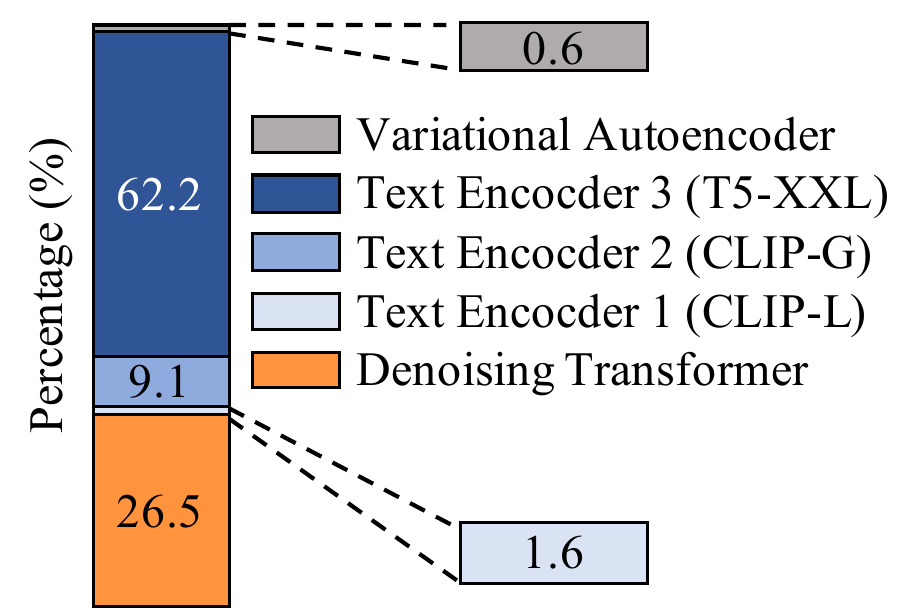}
    \caption{Parameter ratio.}
    \label{fig:FLOPs_ratio}
  \end{subfigure}
  \caption{
\textbf{(a)} FLOPs distribution during image generation in Stable Diffusion 3 (SD3)~\cite{esser2024scaling}. \textbf{(b)} Parameter distribution across modules in SD3. The text encoders contributes less than 0.5\% to the overall FLOPs but account for over 70\% of the total model parameters. For VAE, only the decoder was considered.}
  \label{fig:big_text_encoders}
\end{figure}

To address these issues, research suggests enhancing the efficiency of the T2I diffusion model through strategies such as knowledge distillation (KD)~\cite{castells2024edgefusion, li2024snapfusion, song2024multi, kim2025bk, zhao2025mobilediffusion}, which transfers knowledge from larger models to smaller ones; pruning~\cite{castells2024ld, ganjdanesh2024not, lee2024dit, wang2024patch}, which eliminates superfluous weights; and quantization~\cite{li2023q, he2024ptqd, ryu2025dgq}, which reduces precision by utilizing fewer bits. While effective, these methods target the denoising module. As shown in Fig.~\ref{fig:big_text_encoders}, the text encoders account for over 70\% of the total parameters, but only 0.5\% of floating-point operations (FLOPs), causing disproportionate memory usage. Despite this imbalance, efforts to reduce the text encoder size have been limited.

Large language models (LLMs) also face similar challenges, where model sizes result in significant computational and memory overhead, hindering their practical use. To address this problem, studies such as KD~\cite{hsieh2023distilling, huang2023towards, ko2024distillm}, quantization~\cite{xiao2023smoothquant, ashkboos2024quarot, lin2024awq}, and pruning~\cite{sun2023simple, ashkboos2024slicegpt, men2024shortgpt, song2024sleb, yang2024laco, zhang2024finercut} have been proposed. While KD reduces the size of the model, it requires costly training. Quantization reduces memory usage by reducing precision, but it requires specific hardware support. In contrast, pruning offers parameter reduction with minimal performance loss, making it an efficient solution.

Among pruning techniques, structured pruning has been actively studied to remove rows and columns~\cite{van2023llm, ashkboos2024slicegpt} of the model weights, or entire layers or blocks~\cite{gromov2024unreasonable, men2024shortgpt, yang2024laco, zhang2024finercut} of the model to reduce the size of the model and improve the inference speed. However, these methods are designed for autoregressive LLMs and face challenges when applied to T2I diffusion models, limiting their effectiveness in this context.

We propose \textbf{Sk}ip and \textbf{r}euse the laye\textbf{r}s (Skrr), a blockwise pruning technique for T2I diffusion models. Skrr effectively reduces text encoder size, alleviating memory overhead while preserving image quality and text alignment. Skrr involves two primary stages: the \textit{Skip} to detect layers for pruning, followed by the \textit{Re-use} to recycle the remaining layers to mitigate performance degradation. To the best of our knowledge, this is the first work to tackle the challenge of constructing a lightweight text encoder for T2I tasks.

In the Skip phase, sub-blocks of the text encoder transformer are pruned using a T2I diffusion-tailored discrepancy metric to align dense and pruned models. Prior methods greedily remove blocks to reduce computational costs, but often overlook block interactions, leading to suboptimal pruning. To mitigate this, we propose a beam search~\cite{freitag2017beam}-based approach that explores multiple pruning paths simultaneously, balancing the performance of exhaustive and efficiency of greedy strategies. In the Re-use phase, we assess the discrepancy from reusing adjacent un-skipped blocks to identify those that can be effectively reutilized. Additionally, we provide theoretical support showing that Re-use can enhance performance beyond mere skipping. To improve discrepancy measurement in both phases, we employ a projection module identical to the one used for conditioning text embeddings in the denoising module.

We conducted comprehensive experiments across various metrics, sparsity levels, and diffusion models to thoroughly evaluate the T2I performance of compressed text encoders. The results indicate that Skrr surpasses current autoregressive LLM-targeted pruning techniques on image fidelity and text-image alignment in high sparsity ($\mathbf{>40\%}$).

Our contributions can be summarized as follows.
\begin{itemize}
    \vspace{-0.5em}
    \item We propose \textbf{Skrr}, an effective layer pruning method for the text encoder in T2I diffusion models.
    \vspace{-0.5em}
    \item We present \textit{Skip}, a pruning approach for lightweight T2I diffusion models, and \textit{Re-use}, a method to restore T2I performance by leveraging the remaining layers, supported by theoretical analysis.
    \vspace{-0.5em}
    \item Skrr achieves state-of-the-art blockwise pruning for T2I synthesis, improving GenEval scores by up to 20.4\% at high sparsity over 40\%.
\end{itemize}

\section{Related Works}
\label{related_works}
\subsection{Efficient diffusion model}
As diffusion generative models scale, T2I synthesis has achieved impressive results in generating high-fidelity, text-aligned images. However, this advancement comes with significant computational and memory overhead. Previous research has primarily focused on optimizing the efficiency of the denoising module through methods such as knowledge distillation~\cite{li2024snapfusion, song2024multi, castells2024edgefusion, zhao2025mobilediffusion, kim2025bk}, pruning model weights~\cite{fang2023structural, ganjdanesh2024not, wang2024patch, castells2024ld, lee2024dit}, and quantization of weights to lower precision bits~\cite{li2023q, he2024ptqd, li2024svdqunat, wang2024quest, ryu2025dgq}. While these approaches effectively reduce the computational and memory costs of the pipeline, they overlook the substantial memory burden imposed by the text encoder, which remains largely underexplored. To address this gap, we propose a targeted pruning strategy for the text encoder of a T2I diffusion model, which allows more memory efficient T2I synthesis with comparable performance.

\subsection{Blockwise pruning for LLMs}
Although LLMs show promising performance in various tasks, their size often limits practical deployment. To address this problem, model compression techniques have been developed, with pruning emerging as a promising solution due to its ability to reduce the parameter count with minimal retraining. Notably, blockwise pruning of transformers~\cite{men2024shortgpt, yang2024laco, zhang2024finercut} effectively reduces parameters while preserving performance.

ShortGPT~\cite{men2024shortgpt} proposed a method to prune blocks to the desired sparsity by removing them one by one and assigning a Block Influence (BI) score based on changes in cosine similarity and pruning those with lower BI first. LaCo~\cite{yang2024laco} introduced a strategy to reduce the size of LLMs by merging the weights of adjacent transformer blocks, effectively compressing the model. FinerCut~\cite{zhang2024finercut} refined this approach by pruning sub-blocks, composed of Multi-Head Attention (MHA) and Feed-Forward Network (FFN) layers with normalization, in a fine-grained manner. It sequentially pruned sub-blocks while partially considering interactions between blocks, 
reducing the performance gap with the dense model.

\begin{figure*}[!t]
  \centering
  \begin{subfigure}{0.49\linewidth}
    \includegraphics[width=1\linewidth]{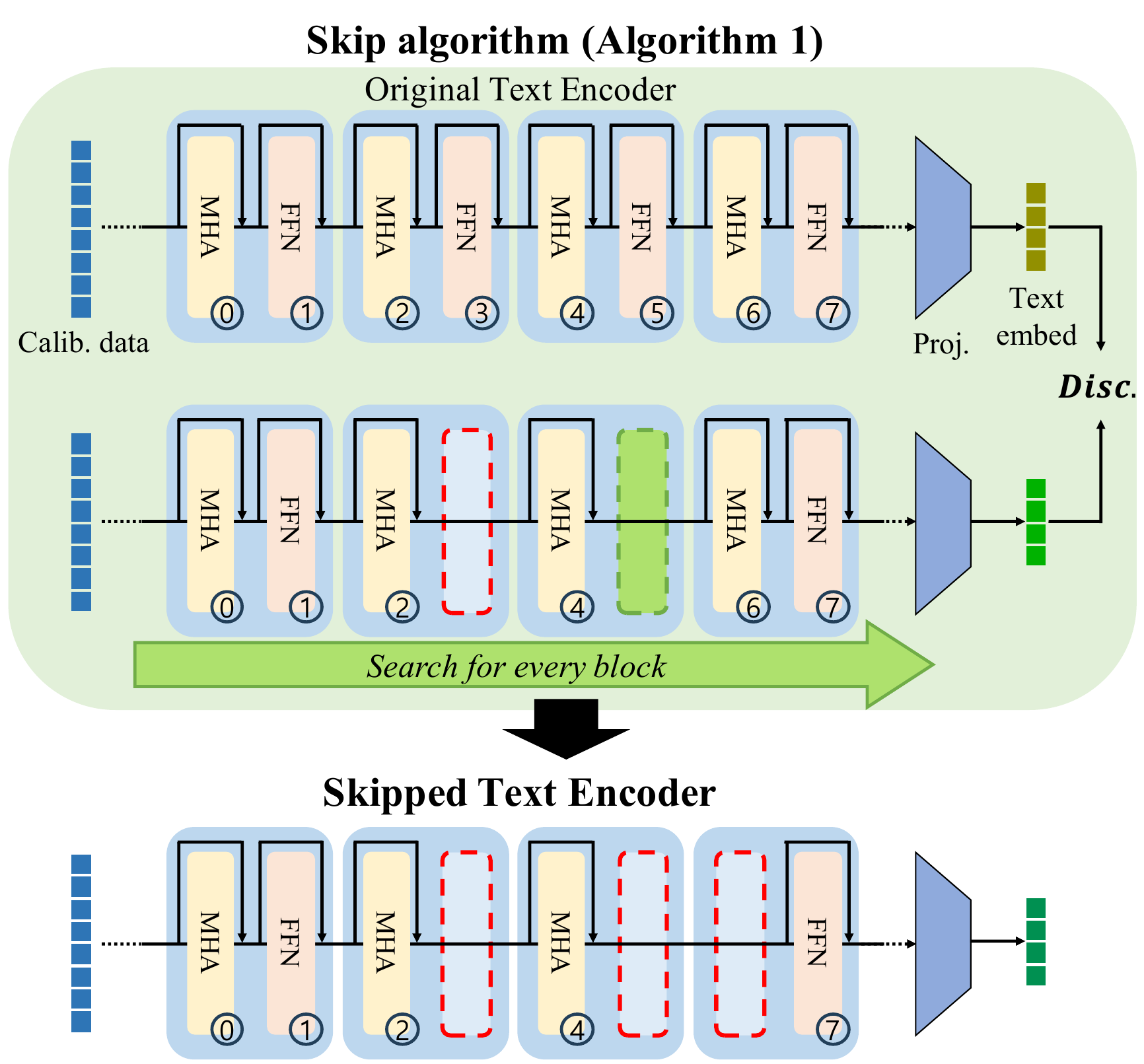}
    \caption{Illustration of Skip phase.}
    \label{fig:overall_framework_skip}
  \end{subfigure}
  \hfill
  \begin{subfigure}{0.49\linewidth}
    \includegraphics[width=1\linewidth]{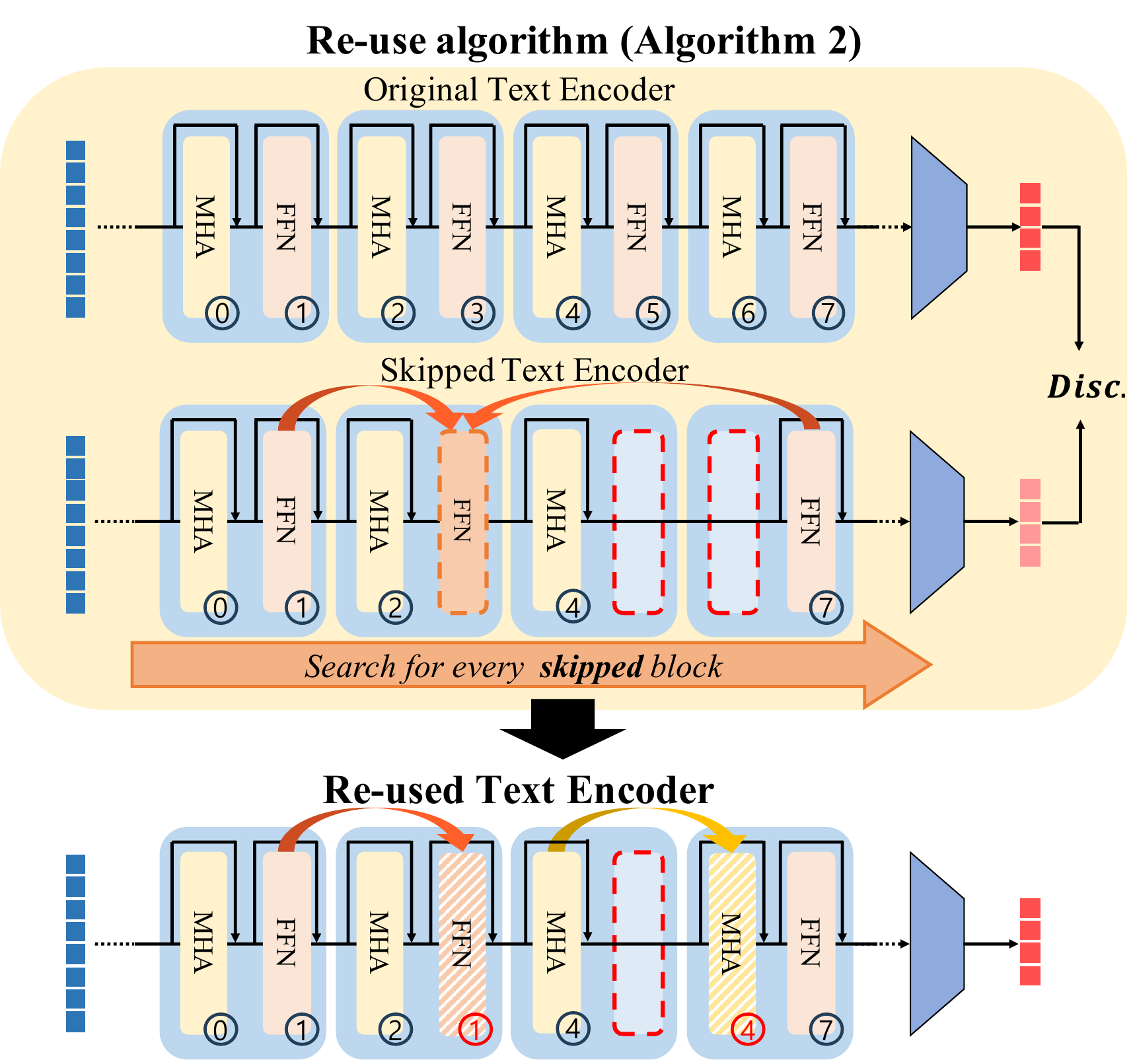}
    \caption{Illustration of Re-use phase.}
    \label{fig:overall_framework_reuse}
  \end{subfigure}
  \vspace{-1em}
  \caption{
The visualization of overall framework of Skrr. \textbf{(a)} shows the \textit{Skip} phase, which repeatedly assesses each sub-block by determining the output discrepancy (Disc.) between the dense and skipped models using a calibration dataset (Calib. data). To account for block interactions, it keeps the top $k$ options with the smallest discrepancies and uses beam search for refined selection. \textbf{(b)} presents the \textit{Re-use} phase, evaluating if recycling remaining block instead of skipped sub-blocks results in a smaller output discrepancy. If so, hidden states are fed back into the chosen layers. This two-phase approach efficiently reduces model size with minimal T2I performance loss.}
  \label{fig:overall_framework}
\vspace{-1em}
\end{figure*}

Despite these advances, text encoder compression in diffusion models remains underexplored. To address this discrepancy, we propose Skrr, a blockwise text encoder pruning approach tailored for T2I diffusion models. We evaluated Skrr against existing LLM-based pruning techniques, demonstrating its effectiveness in reducing the size of the model while maintaining the T2I performance of its dense model.

\section{Method}
\label{method}
Skrr is built around two main parts: \textit{Skip} identifies layers to prune, and \textit{Re-use} selects layers to reuse from the remained ones. 
During the \textit{Skip} phase, each multi-head attention (MHA) and feed-forward network (FFN) sub-block is individually evaluated for its importance using a T2I diffusion-tailored metric. The sub-blocks are then ranked based on their significance. To optimize the pruning process, blocks with low importance are removed sequentially while exploring multiple possible combinations using a beam search-based algorithm. The \textit{Re-use} phase evaluates each layer to reuse a layer based on the metric leveraged in \textit{Skip} phase to the original output, ensuring that important information is conserved, thus reducing performance loss. The overall framework of Skrr is depicted in Fig.~\ref{fig:overall_framework}.

\subsection{Skip Algorithm}
\begin{figure}[!t]
  \centering
  \begin{subfigure}{0.48\linewidth}
    \includegraphics[width=1\linewidth]{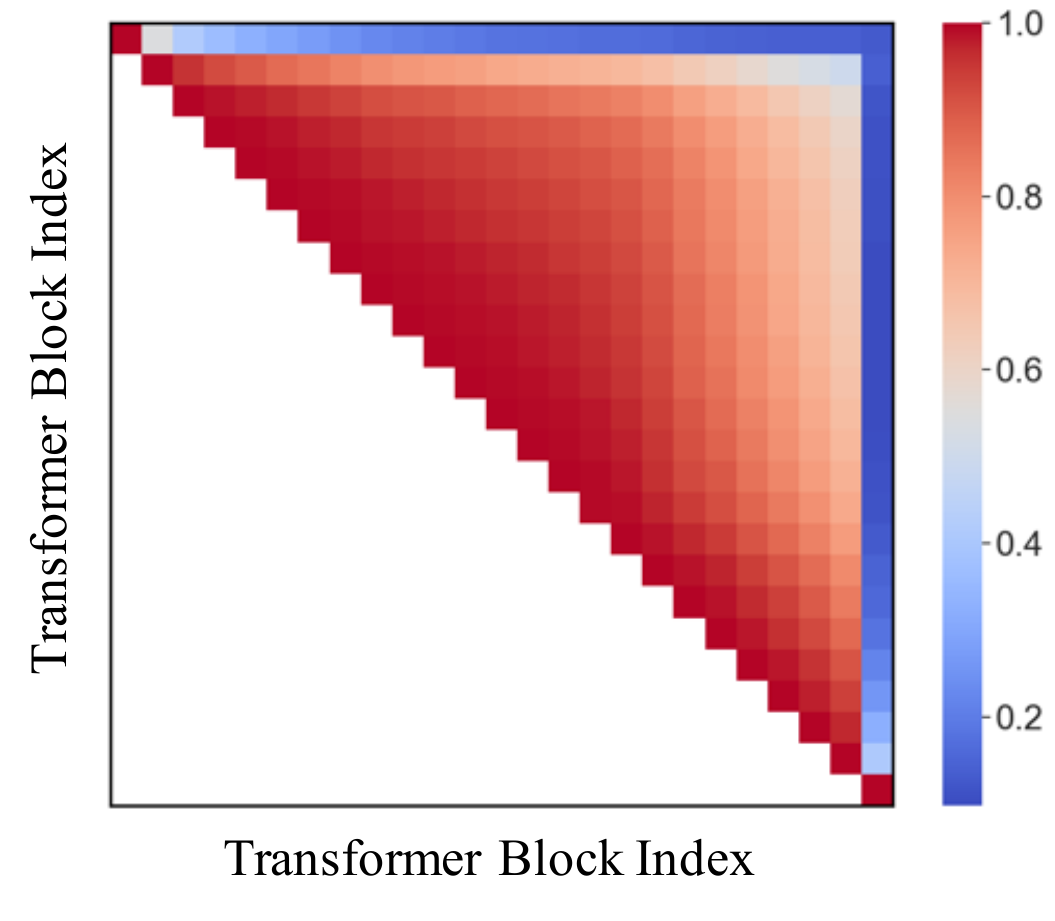}
    \caption{Hidden states similarity.}
    \label{fig:feature_similarity}
  \end{subfigure}
  \begin{subfigure}{0.48\linewidth}
    \includegraphics[width=1\linewidth]{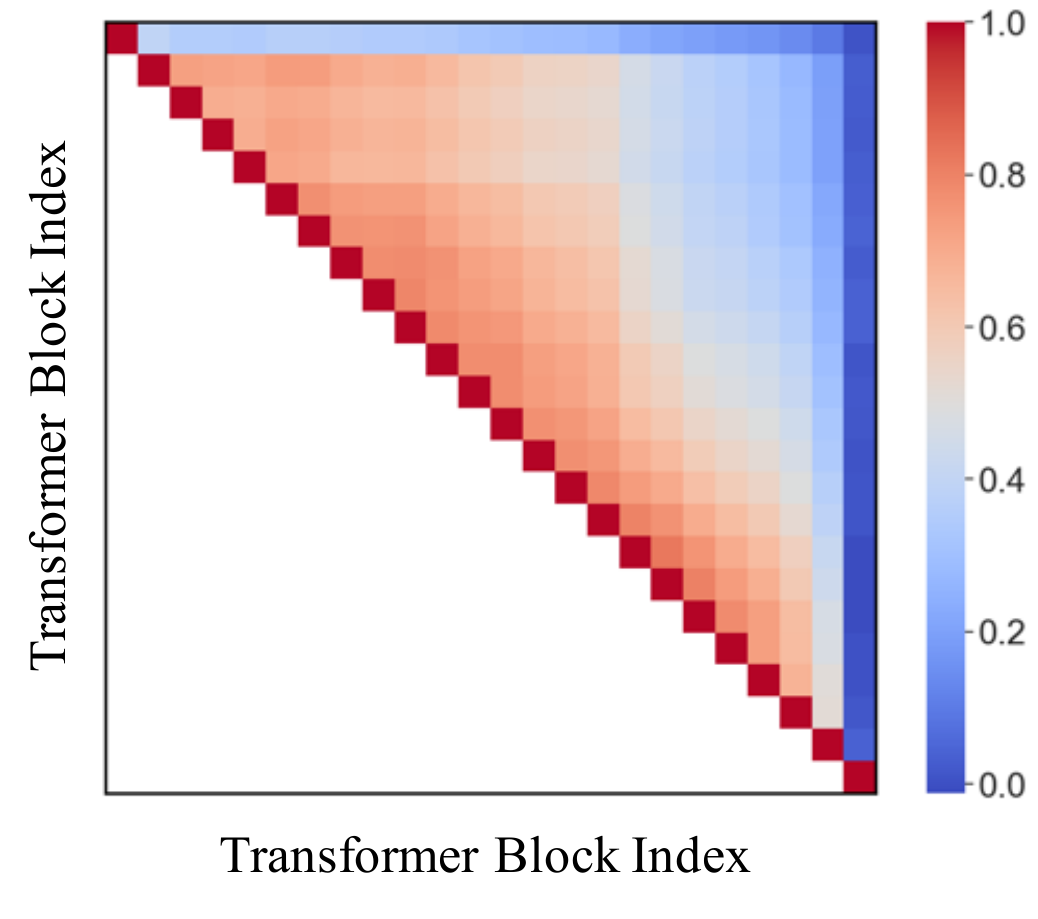}
    \caption{Fixed input similarity.}
    \label{fig:block_similarity}
  \end{subfigure}
  \caption{
\textbf{(a)} The cosine similarity of hidden states in T5 transformer blocks demonstrates progressive variations. The result indicates specific layers could be omitted without serious performance degradation. \textbf{(b)} The cosine similarity of block outputs using fixed inputs. The similarity map reveals redundant role across blocks, suggesting that certain blocks could be replaced by adjacent blocks.}
  \label{fig:similarity}
\end{figure}

\paragraph{Feasibility of skipping blocks.}
Prior work~\cite{men2024shortgpt,yang2024laco,zhang2024finercut} shows transformer blocks can be pruned based on output similarity in LLMs. We extend this analysis to text encoders in diffusion models, especially T5-XXL~\cite{raffel2020exploring}, widely used in T2I models, as shown in Fig.~\ref{fig:feature_similarity}. We observed a high degree of similarity between the hidden states of adjacent blocks. This strong similarity underscores the redundancy in the model and confirms the potential to prune blocks without significantly compromising performance.

\paragraph{Discrepancy metric.}
For effective block removal, it is crucial to evaluate impact of pruning each block on output. A strong metric must be defined to measure text embedding changes of post-removal. Given that text embeddings affect image quality in text-to-image models, choosing the right metric is vital.
Current T2I diffusion models predominantly employ transformer frameworks to synthesize images from input noise and text embeddings. However, text embeddings from the output of the text encoder are not used as is. They undergo alignment through a single linear layer or multi-layer perceptron (MLP), which is represented as:
\begin{equation}\label{eq:projection}
    f = \text{proj}(E(c;\theta_{\text{text.}});\theta_{\text{denoise.}}) ,
\end{equation}
where $c$ is the input prompt, $E(\cdot;\theta_{\text{text.}})$ is the text encoder parameterized with $\theta_{\text{text.}}$, $\text{proj}(\cdot;\theta_{\text{denoise.}})$ is the projection layer in the denoising module for condition vector from the text encoder.
Using the features extracted in this manner, the importance of a block can be evaluated by comparing the similarity between the feature $f_{\text{dense}}$ of the dense model and the feature $f_{\text{skip}}$ of the model that skips (prunes) a block. A commonly used metric in prior studies~\cite{men2024shortgpt, yang2024laco, zhang2024finercut} for measuring discrepancy is cosine similarity, which is formulated as:
\begin{equation}\label{eq:cossim}
    \text{Metric}_1(f_{\text{dense}}, f_{\text{skip}}) = 1 - \frac{f_{\text{dense}} \cdot f_{\text{skip}}}{||f_{\text{dense}}||_2||f_{\text{skip}}||_2}
\end{equation}
Another metric worth considering is the mean-squared error (MSE) between two vectors, which differs from angular metrics by accounting for both the direction (angle) and the magnitude of the vectors. The MSE is formulated as:
\begin{equation}\label{eq:mse}
    \text{Metric}_2(f_{\text{dense}}, f_{\text{skip}}) = \frac{1}{d}\sum_{i=1}^{d}(f_{\text{dense}}^i - f_{\text{skip}}^i)^2 ,
\end{equation}
where $d$ is the dimension of the output feature vectors $f_{\text{dense}}$ and $f_{\text{skip}}$, and $f^i$ represents the $i$-th component of vector $f$. $\text{Metric}_1$ considers only the angle between vectors, while $\text{Metric}_2$ integrates both the angle and magnitude, offering a more comprehensive evaluation of output discrepancies between dense and skipped models~\cite{zhang2024finercut}. Therefore, we chose $\text{Metric}_2$ to assess the discrepancies.

\begin{figure}[!t]
  \centering
  \footnotesize
  \begin{subfigure}[t]{0.32\linewidth}
    \includegraphics[width=1\linewidth]{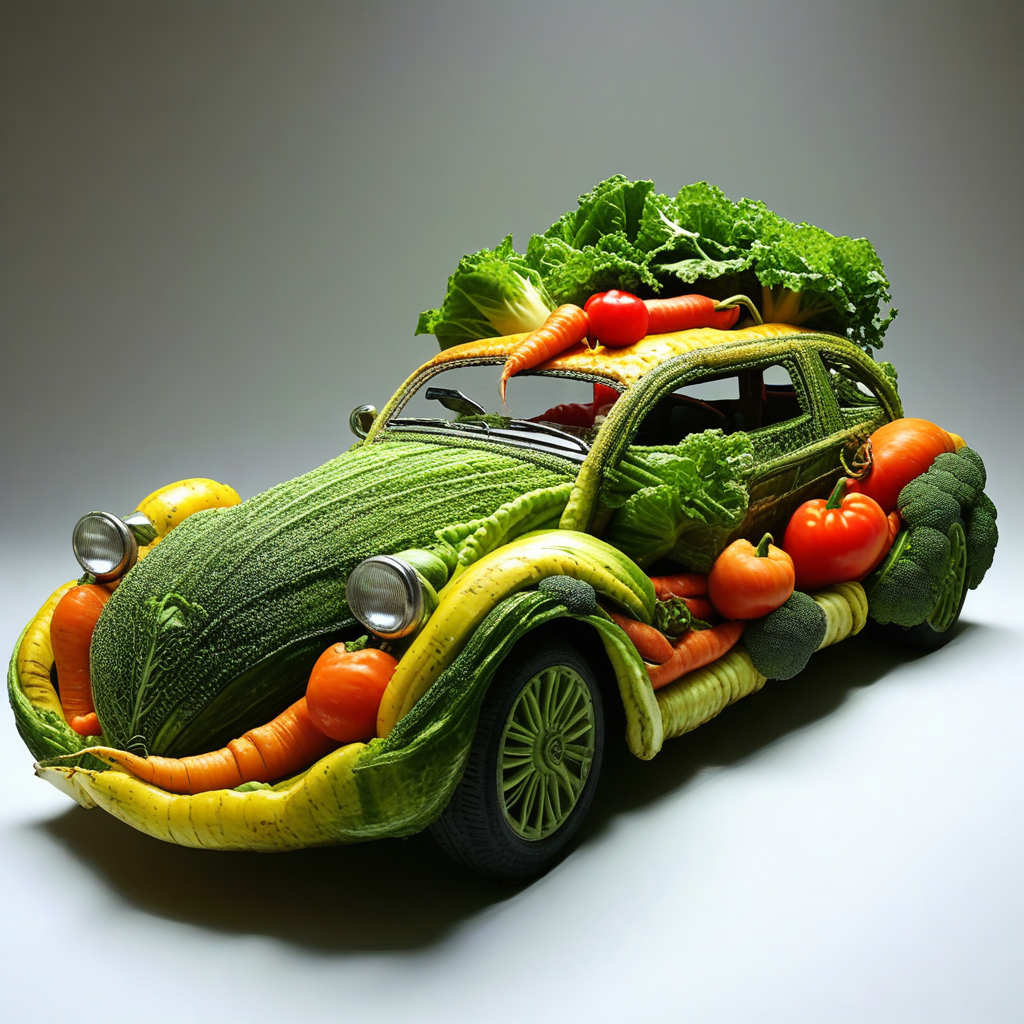}
    \caption{Original image.}
    \label{fig:null_original}
  \end{subfigure}
  \hfill
  \begin{subfigure}[t]{0.32\linewidth}
    \includegraphics[width=1\linewidth]{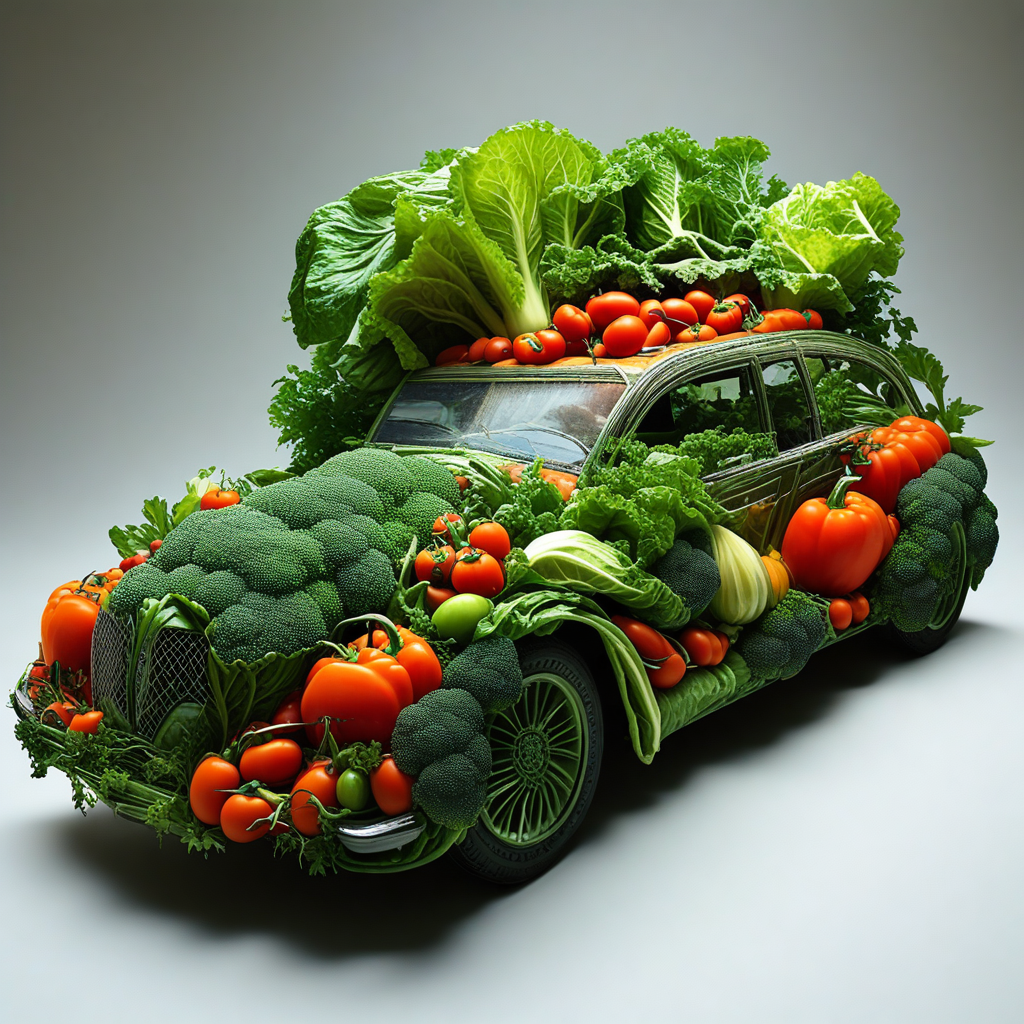}
    \caption{$7^{\text{th}}$, $22^{\text{th}}$ removed.}
    \label{fig:null_normal}
  \end{subfigure}
  \hfill
  \begin{subfigure}[t]{0.32\linewidth}
    \includegraphics[width=1\linewidth]{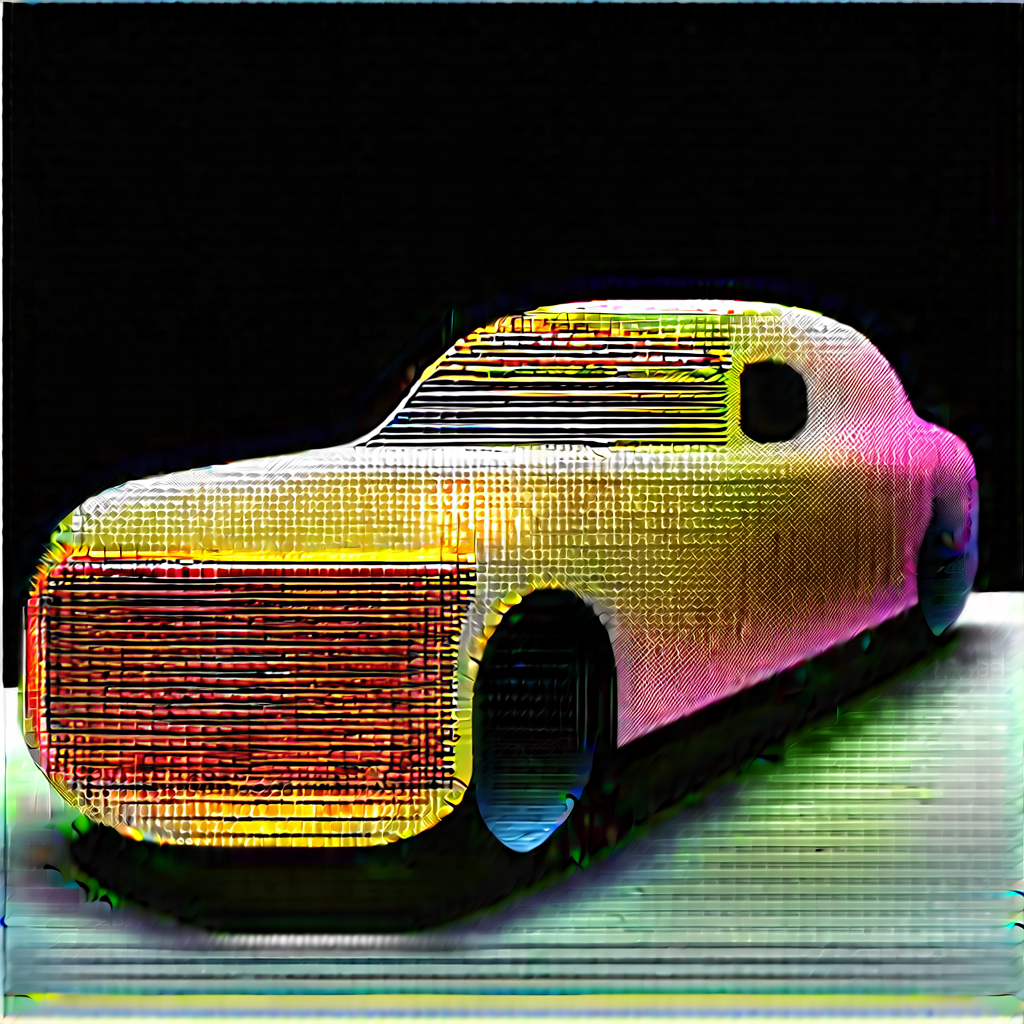}
    \caption{$3^{\text{rd}}$, $5^{\text{th}}$ removed.}
    \label{fig:null_exploded}
  \end{subfigure}
  \caption{
\textbf{(a)} An image is created by the PixArt-$\Sigma$ dense text encoder using the prompt \textit{``A car made out of vegetables.''} with $||f_{\varnothing}||_2=0.03$. For image \textbf{(b)}, the $7^{\text{th}}$ and $22^{\text{th}}$ sub-blocks are excluded, resulting in $\text{Metric}_1 = 0.85$, $\text{Metric}_2 = 0.002$, and $||f_{\varnothing}||_2 = 0.19$. Image \textbf{(c)} is generated by removing the $3^{\text{rd}}$ and $5^{\text{th}}$ sub-blocks, producing $\text{Metric}_1 = 0.89$, $\text{Metric}_2 = 0.04$, and $||f_{\varnothing}||_2 = 3.34$. Despite $\text{Metric}_1$ being higher in (c), the large $||f_{\varnothing}||_2$ value compared to (b) leads to an abnormal image. Notably, $\text{Metric}_2$ more accurately indicates differences in image quality.}
  \label{fig:null_condition}
\end{figure}

\paragraph{Null condition discrepancy.}
In diffusion generative models, guidance is essential to produce high-quality images. A prevalent approach is classifier-free guidance (CFG)~\cite{ho2021classifier}, which improves conditional synthesis by utilizing unconditional scores. This technique derives an unconditional score from a null condition, extrapolates it with the conditional score, and is formulated as follows:
\begin{equation}
    \tilde{\epsilon}(x_t, z) = (1 + w)\epsilon(x_t, f_c) - w \epsilon(x_t, f_{\varnothing}) ,
\end{equation}
where $\epsilon(\cdot,\cdot)$ is the denoising score network, $x_t$ is a noisy sample at timestep $t$, $f_c$ denotes the condition vector from Eq.~\eqref{eq:projection}, $f_{\varnothing}$ represents the null condition vector, and $w$ is the guidance scale. While guidance improves image quality, excessive scaling may cause over-saturation or artifacts. We observed that pruning certain text encoder blocks amplifies its norm by over $100 \times$ compared to the dense version, leading to abnormal images. As illustrated in Fig.~\ref{fig:null_condition}, pruning even two blocks significantly increases $||f_{\varnothing}||_2$, causing abnormalities shown in Fig.~\ref{fig:null_exploded}. Thus, discrepancies in $f_{\varnothing}$ must be considered. Notably, in Fig.~\ref{fig:null_exploded}, $\text{Metric}_1$ failed to assess image quality, while $\text{Metric}_2$ reported smaller values for Fig.~\ref{fig:null_normal}, confirming its reliability and effectiveness.

\paragraph{Beam search-based strategy.}
Most blockwise pruning methods perform (1) rank the importance of the block per layer and pruning in order~\cite{men2024shortgpt}, or (2) perform sequential pruning while re-evaluating blocks~\cite{yang2024laco, zhang2024finercut}. The first method is efficient, but may ignore block interactions and lead to suboptimal results. The second method acknowledges interactions among blocks, yet its greedy approach can still lead to suboptimal results. To address these issues, we propose a novel method akin to beam search, which evaluates multiple pruning paths concurrently to better account for block interactions, achieving more effective pruning without degrading performance.

\paragraph{Algorithm.}
To synthesize the proposed approach, we introduce the Skip algorithm in Algorithm~\ref{alg:block_prune_skip}. Before delving into the specifics, we define two key discrepancy metrics: $D_{f_c}$, derived from $\text{Metric}_2$, and $D_{f_\varnothing}$, representing $\text{Metric}_2$ with null inputs. The algorithm is inspired by a beam search~\cite{freitag2017beam}, iterating over each unskipped sub-block while maintaining the $k$ beams with the smallest sum of $D_{f_c}$and $D_{f_\varnothing}$. This process is repeated from the $k$ beams, iteratively updating them to ensure smaller $D$. Upon traversing all blocks, the algorithm produces a list of Skip indices $\mathcal{S}^*$, effectively capturing inter-block interactions. Finally, the blocks are pruned sequentially according to the $\mathcal{S}^*$ to achieve the desired sparsity.

\begin{algorithm}[t]
\caption{Skip Algorithm}
\label{alg:block_prune_skip}
\begin{algorithmic}[1]
\REQUIRE Calibration dataset $\mathcal{C}$, dense model $\mathcal{M}$, null input $c_{\varnothing}$, number of layers $L$, beam size $k$
\ENSURE Skip index list $\mathcal{S}^*$

\STATE $\mathcal{S} \gets [\,]$
\STATE $\mathcal{B} \gets \{(0,\mathcal{S})\}$,\quad $\mathcal{S}^* \gets \mathcal{S}$

\WHILE{$\exists (D, \mathcal{S}) \in \mathcal{B} \text{ such that } |\mathcal{S}| < L$}
    \STATE $\mathcal{B}_{\mathrm{new}} \gets \emptyset$
    \FOR{each $(D, \mathcal{S}) \in \mathcal{B}$}
        \FOR{each layer index $i \notin \mathcal{S}$}
            \STATE $\mathcal{S}' \gets \text{Append}(\mathcal{S}, i)$
            \STATE $\hat{\mathcal{M}} \gets \text{Prune}(\mathcal{M}, \mathcal{S}')$
            \STATE $D' \gets \text{GetDiscrepancy}(\mathcal{M}, \hat{\mathcal{M}}, \mathcal{C}, c_{\varnothing})$
            \STATE $\mathcal{B}_{\mathrm{new}} \gets \mathcal{B}_{\mathrm{new}} \cup \{(D', \mathcal{S}')\}$
        \ENDFOR
    \ENDFOR
    \STATE Update $\mathcal{B}_{\text{new}}$ to smallest $k$ candidates with $D$
    \STATE $(D^*, \mathcal{S}^*) \gets \arg\min_{(D', \mathcal{S}')} D' \text{ in } \mathcal{B}$
\ENDWHILE
\STATE \textbf{return} $\mathcal{S}^*$
\vspace{-1em}
\end{algorithmic}
\hrulefill

\textbf{Function Definitions:}
\begin{algorithmic}[1]
\STATE \textbf{GetDiscrepancy($\mathcal{M},\hat{\mathcal{M}}, \mathcal{C}, c_{\varnothing}$):}
\STATE \hspace{1em} $D_{f_c} \gets \text{MSE}(\mathcal{M}(\mathcal{C}), \hat{\mathcal{M}}(\mathcal{C}))$
\STATE \hspace{1em} $D_{f_{\varnothing}} \gets \text{MSE}(\mathcal{M}(c_{\varnothing}), \hat{\mathcal{M}}(c_{\varnothing})) $
\STATE \hspace{1em} \textbf{return} $D_{f_c} + D_{f_{\varnothing}}$
\end{algorithmic}
\end{algorithm}
\vspace{-0.5em}

\subsection{Re-use Algorithm}
\paragraph{Feasibility of reusing blocks.}
Despite extensive research on methods such as recurrent networks~\cite{sherstinsky2020fundamentals, gu2023mamba} which loop the output of the network back into the input for efficiency, the practice of reusing specific layers in neural networks by reintegrating hidden states into internal layers remains understudied. We conducted a feasibility study to investigate whether the internal components of the T2I diffusion text encoder serve analogous functions (see Fig.~\ref{fig:block_similarity}). We randomly sampled tokens from the embedding, passed them through each transformer block as a fixed input, and measured the similarity of their output. The results indicate significant similarity between adjacent blocks, suggesting that performance can be restored by reintroducing non-omitted layers into adjacent skipped layers. We also verified the existence of condition for Re-use that achieves a tighter error bound compared to Skip alone. The existence is formalized in Theorem~\ref{thm:feasibility}, which theoretically demonstrates the advantages of incorporating the Re-use phase. To establish this result, we first introduce the following lemma.
\begin{lemma}[Error bound of two transformers]\label{lemma:error_bound}
Let $\mathcal{M}: (x, \theta) \mapsto \mathbb{R}^d$ be an $L$-block transformer with input $x \in \mathbb{R}^d$ and parameter set $\theta = (\theta_1, \dots, \theta_L)$, defined as:
\begin{equation}
    \mathcal{M} = \big((F_L + I) \circ (F_{L-1} + I) \circ \cdots \circ (F_1 + I)\big)
\end{equation}
where $F_i: (z_i, \theta_i) \mapsto \mathbb{R}^d$ is the $i$-th block with parameters $\theta_i$, and \( z_i \in \mathbb{R}^d \).
Assume that \( F_i \) is \( L_i \)-Lipschitz in \( z_i \) and \( M_i \)-Lipschitz in \( \theta_i \). Then, for any two parameter sets \( \theta = (\theta_1, \dots, \theta_L) \) and \( \hat{\theta} = (\hat{\theta}_1, \dots, \hat{\theta}_L) \), the following holds:
\begin{equation}
\begin{split}
&\big\| \mathcal{M}(x; \theta) - \mathcal{M}(x; \hat{\theta}) \big\| \\
&\le \sum_{i=1}^L \bigg( \prod_{k=i+1}^L (1 + L_k) \bigg) M_i \big\| \theta_i - \hat{\theta}_i \big\|\coloneqq U
\end{split}
\end{equation}
\end{lemma}
The proof for Lemma~\ref{lemma:error_bound} is in the Appendix Sec.~\ref{append:proof_lemma}. With the lemma, we can prove the following theorem.
\begin{theorem}[Tighter error bound of Re-use]\label{thm:feasibility}
Under the assumptions of Lemma~\ref{lemma:error_bound}, let $\theta^*_i$ be the parameters of the reused $F_i$. Define $U_{\text{Skip}}$ as the error bound for the compressed model with Skip alone and $U_{\text{Skip, Re-use}}$ as the error bound for the compressed model with Skip and Re-use. If $\|\theta_i - \theta_i^*\| < \|\theta_i\|$,
then the following holds:
\begin{equation}
    U_{\text{Skip, Re-use}} < U_{\text{Skip}}.
\end{equation}
\end{theorem}
This theorem establishes the theoretical feasibility of Re-use by showing a existence of condition under which the error bound becomes tight with its application.
The proof for Theorem~\ref{thm:feasibility} is shown in the Appendix Sec.~\ref{append:proof_thm}.

\begin{algorithm}[t]
\caption{Re-use Algorithm}
\label{alg:block_prune_reuse}
\begin{algorithmic}[1]
\REQUIRE Calibration dataset $\mathcal{C}$, dense model $\mathcal{M}$, null input ${c}_{\varnothing}$, skip indices list $\mathcal{S}$, re-use indices dictionary $\mathcal{R}$
\ENSURE Re-use indices dictionary $\mathcal{R}$
\STATE $\mathcal{R} \gets \varnothing$, $\hat{\mathcal{M}} \gets \text{Prune}(\mathcal{M}, \mathcal{S})$
\STATE $D_{\mathcal{M}} \gets \text{GetDiscrepancy}(\mathcal{M},\hat{\mathcal{M}}, \mathcal{C}, c_{\varnothing})$
\FOR{each $i \in \mathcal{S}$}
    \small %
    \STATE $l \gets \max \{j < i \mid j \notin \mathcal{S}\}, \; r \gets \min \{j > i \mid j \notin \mathcal{S}\}$
    \STATE $\hat{\mathcal{M}}_l \gets \text{Update}(\hat{\mathcal{M}}, \mathcal{R} \cup \{i:l\}$)
    \STATE $\hat{\mathcal{M}}_r \gets \text{Update}(\hat{\mathcal{M}}, \mathcal{R} \cup \{i:r\}$)
    \STATE $D_{\mathcal{M}} \gets \text{GetDiscrepancy}(\mathcal{M},\hat{\mathcal{M}}, \mathcal{C}, c_{\varnothing})$
    \STATE $D_{l} \gets \text{GetDiscrepancy}(\mathcal{M},\hat{\mathcal{M}}_l, \mathcal{C}, c_{\varnothing})$
    \STATE $D_{r} \gets \text{GetDiscrepancy}(\mathcal{M},\hat{\mathcal{M}}_r, \mathcal{C}, c_{\varnothing})$
    \IF{$D_{l} < D_{\mathcal{M}} \land D_{l} < D_{r}$}
        \STATE $\mathcal{R} \gets \mathcal{R} \cup \{i:l\}$
    \ELSIF{$D_{r} < D_{\mathcal{M}} \land D_{r} < D_{l}$}
        \STATE $\mathcal{R} \gets \mathcal{R} \cup \{i:r\}$
    \ENDIF
    \STATE $\hat{\mathcal{M}} \gets \text{Update}(\hat{\mathcal{M}}, \mathcal{R})$
\ENDFOR
\STATE \textbf{return} $\mathcal{R}$
\end{algorithmic}
\end{algorithm}

\begin{table*}[ht]
    \caption{Quantitative comparisons of Skrr with baselines. We compared Skrr with the baselines of ShortGPT, LaCo, and FinerCut under three different sparsity scenarios on PixArt-$\Sigma$. The results show that Skrr reliably maintains image fidelity and performs comparable to the dense model across all given sparsity levels. Unlike ShortGPT and LaCo, FinerCut and Skrr use sub-block pruning, hindering direct sparsity level alignment. Sparsity levels were matched as closely as possible for fair evaluation, reflecting how much the compressed model's parameters differ from the dense encoder. ($\uparrow / \downarrow$ denotes that a higher / lower metric is favorable.)}
    \label{tab:quantitative}
    \centering
    \resizebox{\textwidth}{!}{
    \begin{tabular}{cccccccccccc}
        \toprule
        \multirow{2}{*}{Method} & Sparsity & \multirow{2}{*}{FID $\downarrow$} & \multirow{2}{*}{CLIP $\uparrow$} & \multirow{2}{*}{DreamSim $\uparrow$} & \multicolumn{7}{c}{GenEval $\uparrow$}    \\
        \cmidrule(lr){6-12}
        & (\%) & & & & Single & Two & Count. & Colors & Pos. & Color attr. & Overall \\
        \midrule
         Dense & 0.0 & 22.89 & 0.314 & 1.0 & 0.988 & 0.616 & 0.475 & 0.795 & 0.108 & 0.255 & 0.539 \\
        \cmidrule(lr){1-12}
         \multirow{3}{*}{ShortGPT}  & 24.3 & 24.96 & 0.309 &  \cellcolor{tabthird}0.753  & \cellcolor{tabthird}0.944 & \cellcolor{tabthird}0.381 & \cellcolor{tabfirst}0.431 & \cellcolor{tabthird}0.715 & 0.033 & 0.083 & \cellcolor{tabthird}0.431 \\
                                     & 32.4 & 27.28 & 0.294 & 0.651 & \cellcolor{tabthird}0.834 & 0.197 & \cellcolor{tabthird}0.291 & 0.537 & \cellcolor{tabthird}0.048 & 0.038 & 0.324 \\
                                     & 40.5 & 55.26 & 0.215 & 0.357 & 0.306 & 0.025 & 0.090 & 0.100 & 0.0 & 0.0 & 0.087 \\
        \hline
         \multirow{3}{*}{LaCo} & 24.3 & \cellcolor{tabfirst}19.45 & \cellcolor{tabthird}0.311 & 0.726 & 0.909 & 0.336 & \cellcolor{tabthird}0.394 & 0.713 & \cellcolor{tabthird}0.065 & \cellcolor{tabsecond}0.128 & 0.424 \\
                              & 32.4 & \cellcolor{tabthird}24.70 & \cellcolor{tabthird}0.303 & \cellcolor{tabthird}0.677 & 0.781 & \cellcolor{tabthird}0.227 & 0.250 & \cellcolor{tabthird}0.606 & 0.043 & \cellcolor{tabthird}0.040 & \cellcolor{tabthird}0.325 \\
                               & 40.5 & \cellcolor{tabthird}21.60 & \cellcolor{tabthird}0.291 & \cellcolor{tabthird}0.620 & \cellcolor{tabthird}0.784 & \cellcolor{tabthird}0.162 & \cellcolor{tabthird}0.150 & \cellcolor{tabthird}0.489 & \cellcolor{tabthird}0.030 & \cellcolor{tabthird}0.033 & \cellcolor{tabthird}0.275 \\
        \hline
         \multirow{3}{*}{FinerCut} & 26.3 & \cellcolor{tabthird}20.66 & \cellcolor{tabsecond}0.313 & \cellcolor{tabsecond}0.798 & \cellcolor{tabsecond}0.947 & \cellcolor{tabfirst}0.465 & 0.394 & \cellcolor{tabsecond}0.737 & \cellcolor{tabfirst}0.103 & \cellcolor{tabthird}0.105 & \cellcolor{tabsecond}0.458  \\
                                    & 32.2 & \cellcolor{tabsecond}20.49 & \cellcolor{tabsecond}0.313 & \cellcolor{tabsecond}0.771 & \cellcolor{tabsecond}0.903 & \cellcolor{tabfirst}0.409 & \cellcolor{tabsecond}0.344 & \cellcolor{tabsecond}0.697 & \cellcolor{tabsecond}0.078 & \cellcolor{tabfirst}0.128 & \cellcolor{tabsecond}0.426 \\
                                   & 41.7 & \cellcolor{tabsecond}20.36 & \cellcolor{tabsecond}0.308 & \cellcolor{tabsecond}0.731 & \cellcolor{tabsecond}0.841 & \cellcolor{tabsecond}0.306 & \cellcolor{tabsecond}0.306 & \cellcolor{tabsecond}0.628 & \cellcolor{tabsecond}0.050 & \cellcolor{tabfirst}0.073 & \cellcolor{tabsecond}0.367\\
        \hline
        \multirow{3}{*}{\textbf{Skrr (Ours)}} & 27.0 & \cellcolor{tabsecond}20.15 & \cellcolor{tabfirst}0.315 & \cellcolor{tabfirst}0.800 & \cellcolor{tabfirst}0.956 & \cellcolor{tabsecond}0.434 & \cellcolor{tabsecond}0.425 & \cellcolor{tabfirst}0.763 & \cellcolor{tabsecond}0.095 & \cellcolor{tabfirst}0.145 & \cellcolor{tabfirst}0.471 \\
                                               & 32.4 & \cellcolor{tabfirst}20.19 & \cellcolor{tabfirst}0.313 & \cellcolor{tabfirst}0.775 & \cellcolor{tabfirst}0.928 & \cellcolor{tabsecond}0.397 & \cellcolor{tabfirst}0.413 & \cellcolor{tabfirst}0.774 & \cellcolor{tabfirst}0.100 & \cellcolor{tabsecond}0.118 & \cellcolor{tabfirst}0.455 \\
                                               & 41.9 & \cellcolor{tabfirst}19.93 & \cellcolor{tabfirst}0.312 & \cellcolor{tabfirst}0.741 & \cellcolor{tabfirst}0.913 & \cellcolor{tabfirst}0.410 & \cellcolor{tabfirst}0.450 & \cellcolor{tabfirst}0.755 & \cellcolor{tabfirst}0.055 & \cellcolor{tabsecond}0.068 & \cellcolor{tabfirst}0.442\\
        \bottomrule
    \end{tabular}
    }
\end{table*}

\paragraph{Algorithm.}
The Re-use algorithm provided in Algorithm~\ref{alg:block_prune_reuse} enhances performance of pruned models by reintroducing adjacent layers for skipped layers. Starting with pruning based on skip indices $\mathcal{S}$, it evaluates each skipped block for reuse with discrepancy score $D$ between pruned and dense model outputs under three configurations: current state, reuse of the previous sub-block, and reuse of the subsequent sub-block. The configuration with the smallest $D$ is selected, ensuring alignment with the dense model. MHA and FFN sub-blocks are reused as their respective types, maintaining consistency and functionality. The sub-block with the lowest discrepancy is reintroduced, iteratively updating the dictionary for Re-use $\mathcal{R}$ until all skipped layers are evaluated. This results in a refined dictionary $\mathcal{R}$, allowing efficient compression while maintaining performance.

\section{Experiments}
\label{experiments}
\paragraph{Baselines.}
To evaluate the performance of Skrr, we compared multiple LLM-based blockwise pruning techniques for T2I synthesis, including ShortGPT~\cite{men2024shortgpt}, LaCo~\cite{yang2024laco} and FinerCut~\cite{zhang2024finercut}. Pruning in diffusion pipelines specifically targets the T5-XXL~\cite{raffel2020exploring}, which constitutes most of the parameter size. Detailed configurations and implementation specifics are available in Appendix Sec.~\ref{append:baselines} and Sec.~\ref{append:diff_and_text}.

\paragraph{Dataset.}
Most blockwise pruning methods rely on a calibration dataset to identify and remove less influential blocks. To this end, we constructed a calibration set by sampling 1k text prompts from the CC12M~\cite{changpinyo2021conceptual}, specifically tailored for T2I synthesis. The detailed configuration and example are provided in Appendix Sec.~\ref{append:calibration}.

\subsection{Quantitative results}
\label{quanti}

\paragraph{Metrics.}
We evaluated the performance of the Skrr and baseline models with four metrics: Fréchet Inception Distance (FID)~\cite{heusel2017gans}, CLIP~\cite{radford2021learning} score and DreamSim~\cite{fu2023dreamsim} score with MS-COCO~\cite{lin2014microsoft} 30k validation set, alongside GenEval~\cite{ghosh2024geneval}. FID measures the real versus generated image similarity using Inception-v3 features~\cite{szegedy2016rethinking}. The CLIP score measures the semantic alignment of the image-text. DreamSim assesses the composition and color similarity in pruned text encoders versus the original. GenEval measures pruned text encoders' effects on T2I synthesis across six dimensions: single object, two objects, counting, color accuracy, positional alignment, and color attribution. See Appendix Sec.~\ref{append:metrics} for details.

\paragraph{T2I synthesis performance comparison.}
We evaluated the T2I synthesis performance of Skrr against various baselines and metrics, as shown in Table~\ref{tab:quantitative}.  ShortGPT maintains performance at low sparsity, but its performance deteriorates rapidly as sparsity increases. A similar pattern is observed for LaCo. FinerCut exhibits a more gradual decline in performance compared to other baselines, but the generated images still show a significant drop in GenEval scores. In contrast, Skrr demonstrates performance comparable to dense models in all the sparsity levels, achieving high fidelity while preserving strong text alignment metrics especially in high-sparsity settings. Interestingly, in some cases, the FID score for compressed models improves, while DreamSim, CLIP, and GenEval scores degrade. This indicates that compressing text encoders preserves or boosts image quality, while text alignment declines. We provide further detailed analysis in Sec.~\ref{discussion}.

\begin{figure*}[!t]
    \centering
    \includegraphics[width=1\linewidth]{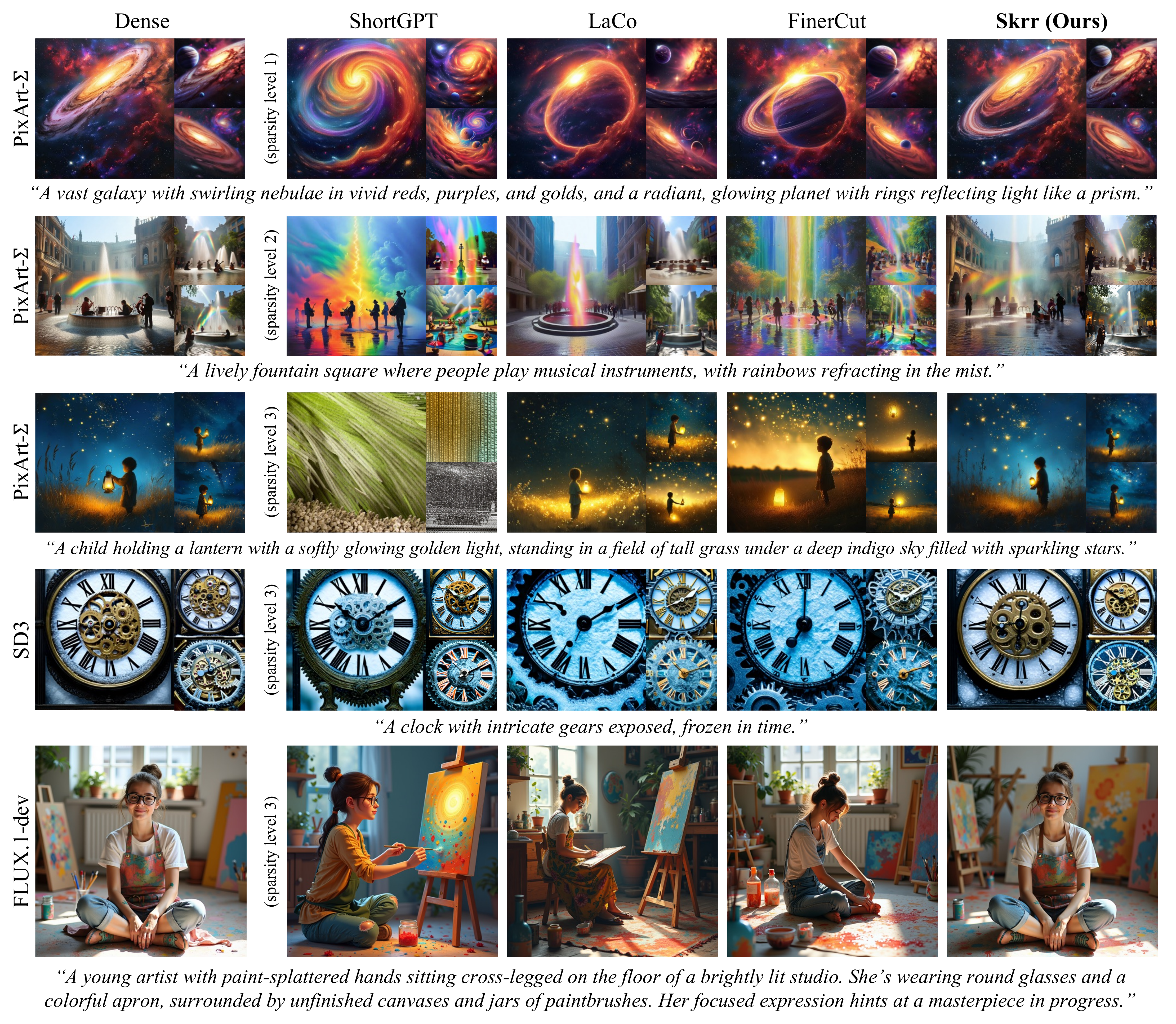}
    \caption{Comparison of images generated with baseline and Skrr-compressed text encoders across PixArt-$\Sigma$, Stable Diffusion 3 (SD3), and FLUX.1-dev. At low sparsity (level 1--24.3\% for ShortGPT and Laco, 26.3\% for FinerCut, and 27.0\% for Skrr), both methods perform comparably to dense models, but Skrr outperforms at higher sparsity(level 2--32.4\% for ShortGPT and Laco, 32.2\% for FinerCut, and 32.4\% for Skrr, level 3--40.5\% for ShortGPT and Laco, 41.7\% for FinerCut, and 41.9\% for Skrr), maintaining alignment to dense model and preserving details in the prompt such as \textit{``glasses''}, \textit{``colorful apron''}, and \textit{``paint-splattered hands''}, where baseline methods fail.}
    \label{fig:qualitative_result}
\end{figure*}

\paragraph{Computational cost.}
To evaluate the efficiency of Skrr, we compared the computational cost of dense and pruned models. We measured the number of parameters, memory usage, and total FLOPs within the PixArt-$\Sigma$ pipeline, with the model precision standardized to Bfloat16. As shown in Table~\ref{tab:computation}, Skrr significantly reduces the parameters and memory usage similar to other baselines compared to dense model. While its FLOPs are slightly higher than the other baselines, this is negligible, since the text encoder accounts for only a small portion (0.6\%) of the pipeline's total FLOPs.

\begin{table}[]
    \caption{Number of parameters (Param.), memory usage (Mem.), and FLOPs were evaluated on PixArt-$\Sigma$ across pruning methods, considering the entire pipeline to analyze each strategy's impact on computational cost. All metrics were measured at the maximum sparsity achieved by each pruning method.}
    \vspace{8pt}
    \label{tab:computation}
    \centering
    \resizebox{\linewidth}{!}{
    \begin{tabular}{cccccc}
    \toprule
    Method & Sparsity (\%) & Param. (B) & Mem. (GB) & TFLOPs \\
    \midrule
    Dense & 0.0 &5.42 & 10.18 & 91.94 \\
    \midrule
    ShortGPT & 40.5 & 3.49 & 6.59 & 91.79 \\
    FinerCut & 41.7 & 3.43 & 6.48 & 91.74 \\
    \textbf{Skrr (Ours)} & 41.9 & 3.43 & 6.46 & 91.90 \\
    \bottomrule
    \end{tabular}
}
\end{table}

\subsection{Qualitative results}
We present qualitative results that demonstrate the performance of T2I synthesis using Skrr-compressed text encoders, extending the experiments to SD3 and FLUX, state-of-the-art diffusion models. As shown in Fig.~\ref{fig:qualitative_result}, ShortGPT achieves satisfactory image quality at low sparsity, but diverges significantly at higher sparsity levels, failing to align with input text at sparsity over 40\% (level 3). LaCo and FinerCut maintain image fidelity across sparsity levels but show reduced alignment with the dense model. In contrast, Skrr consistently preserves both image quality and alignment, closely resembling the outputs of dense models.
For SD3 and FLUX, which use multiple text encoders, image fidelity remains intact at higher sparsity levels, but similarity to dense encoder output decreases. These results highlight Skrr's robustness in preserving image quality and alignment to original image while exhibiting consistent behavior across models under varying sparsity conditions.

\subsection{Ablation study}
We conducted an ablation study to evaluate the contribution of each component in the Skrr framework. Specifically, we analyzed the effectiveness of Re-use, highlighting its role in minimizing performance degradation from Skip. We also examined the size of the beam, demonstrating its effectiveness in block selection. The experiments were carried out on PixArt-$\Sigma$ and a subset of the MS-COCO~\cite{lin2014microsoft} validation dataset with the highest sparsity. We provide additional ablation study in Appendix Sec.~\ref{append:ablation}.

\begin{figure}[!t]
  \centering
  \includegraphics[width=1\linewidth]{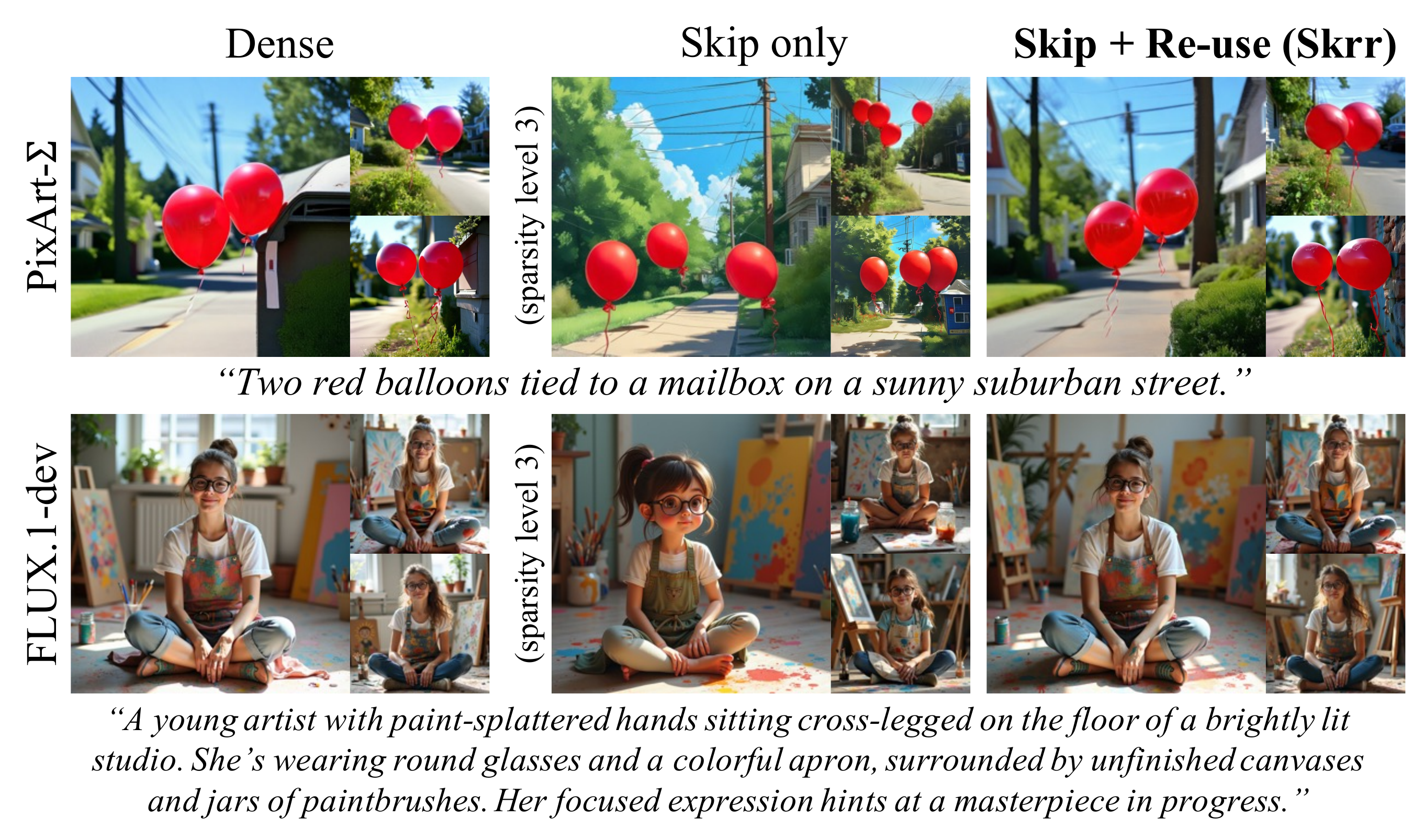}
  \vspace{-2em}
  \caption{Ablation study on Re-use. Without Re-use, Skip alone leads to images that often misalign with the prompt, while Re-use ensures more faithful adherence to the prompt.}
  \label{fig:ablation_reuse}
\vspace{-1em}
\end{figure}

\paragraph{Influence of Re-use.}
The Re-use phase addresses performance degradation from the Skip phase, optimizing within memory constraints. Its impact was assessed by comparing model performance with and without Re-use. As shown in Fig.~\ref{fig:ablation_reuse}, Skip alone preserves high fidelity but may fail to accurately reflect the text prompt. In contrast, incorporating Re-use produces images closely resembling those from the dense model, ensuring better alignment with the prompt.

\paragraph{Influence of beam search.}
We performed an ablation study that evaluated the effect of the beam search at different values of $k$, as shown in Table~\ref{tab:ablation_beam}. As $k$ increases, performance initially improves and then decreases. This trend is consistent with previous findings~\cite{cohen2019empirical, he2023empirical}, which observed similar behavior in LLM decoding in which performance increases and then deteriorates as the size of the beam increases. Based on this observation, we selected the optimal beam size $k=3$.

\begin{table}[t]
    \caption{T2I performance with various beam size $k$. While larger $k$ values incur higher computational costs for the candidate search, they enhance T2I performance. ($k=1$ is a greedy approach.)}
    \label{tab:ablation_beam}
    \centering
    \resizebox{\linewidth}{!}{
    \begin{tabular}{ccccccc}
    \toprule
    Sparsity & \multirow{2}{*}{$k$} & \multirow{2}{*}{CLIP} & \multirow{2}{*}{DreamSim} & \multicolumn{3}{c}{GenEval} \\
    \cmidrule(lr){5-7}
    (\%)& & & & Single. & Count. & Colors. \\
    \midrule
    41.9 & 1 & 0.310 & 0.737 & 0.900 & 0.372 & 0.739 \\
    41.9 & 2 & 0.312 & 0.757 & 0.912 & 0.450 & 0.731 \\
    41.9 & 3 & 0.313 & 0.746 & 0.912 & 0.450 & 0.755 \\
    40.7 & 4 & 0.310 & 0.739 & 0.925 & 0.328 & 0.707 \\
    \bottomrule
    \end{tabular}
}
\vspace{-1em}
\end{table}

\section{Discussion}
\label{discussion}
Previously, we noted better FID results with T2I synthesis when using a text encoder with skipped layers. Beyond CFG, other guidance methods in diffusion models include perturbed attention guidance (PAG)~\cite{ahn2025self} and autoguidance~\cite{karras2024guiding}, both of which approximate the unconditional score by modifications of denoising networks. We propose that layer skipping or merging affects the null text embedding similar to these methods. To verify this hypothesis, we perturbed $f_{\varnothing}$, as described below.
\begin{equation}
    \hat{f}_{\varnothing} = \lambda z + f_{\varnothing} ,\quad z \sim \mathcal{N}(0,I),
\end{equation}
where $z$ is a random vector sampled from normal distribution and $\lambda$ is small scalar value. The FID and CLIP scores of the original model and the unconditional output with small perturbations applied to $f_{\varnothing}$ were measured on the MS-COCO 30k dataset. The results show that the FID decreases, indicating that the fidelity improved when perturbations are introduced to the unconditional feature $f_{\varnothing}$. Detailed results and configurations are provided in Appendix Sec.~\ref{append:perturb}.

\section{Conclusion}
\label{conclusion}
In this paper, we introduce Skip and Re-use Layers (Skrr), an effective compression method for the text encoder in text-to-image (T2I) diffusion models. Skrr integrates three key components: (1) a pruning metric based on the Skrr dot product to identify redundant sub-blocks, (2) a beam search-based algorithm to account for interactions between transformer blocks during pruning, and (3) a re-use mechanism that mitigates performance degradation by leveraging the remaining layers to recover lost capacity from skipped blocks with theoretical supports. Extensive experiments demonstrate that Skrr consistently outperforms existing blockwise pruning techniques for text encoder compression in image synthesis tasks, achieving qualitatively and quantitatively superior results. Additionally, our analysis reveals that pruning or merging layers not only reduces model complexity but can also enhance certain aspects of performance. We further analyze these improvements from the perspective of model guidance, offering insights into how structural adjustments contribute to more effective T2I generation.


\section*{Acknowledgements}
This work was supported in part by Institute of Information \& communications Technology Planning \& Evaluation (IITP) grant funded by the Korea government(MSIT) [NO.RS-2021-II211343, Artificial Intelligence Graduate School Program (Seoul National University)] and the National Research Foundation of Korea(NRF) grant funded by the Korea government(MSIT) (No. NRF-2022M3C1A309202211). Also, the authors acknowledged the financial support from the BK21 FOUR program of the Education and Research Program for Future ICT Pioneers, Seoul National University.

\bibliography{main}
\bibliographystyle{icml2025}

\newpage
\appendix
\renewcommand{\thefigure}{A\arabic{figure}}
\renewcommand{\thetable}{A\arabic{table}}
\renewcommand{\theequation}{A\arabic{equation}}
\setcounter{figure}{0}
\setcounter{table}{0}
\setcounter{equation}{0}
\onecolumn
\section{Proofs}\label{append:proof}
\subsection{Proof of Lemma~\ref{lemma:error_bound}}\label{append:proof_lemma}
\renewcommand{\thetheorem}{3.1}
\begin{lemma}[Error bound of two transformers] Let $\mathcal{M}: (x, \theta) \mapsto \mathbb{R}^d$ be an $L$-block transformer with input $x \in \mathbb{R}^d$ and parameters $\theta = (\theta_1, \dots, \theta_L)$, defined as:
\begin{equation}\label{append:proof_transformer}
    \mathcal{M} = \big((F_L + I) \circ (F_{L-1} + I) \circ \cdots \circ (F_1 + I)\big),
\end{equation}
where $F_i: (z_i, \theta_i) \mapsto z_{i+1}$ is the $i$-th block with parameters $\theta_i$, and \( z_{i+1} \in \mathbb{R}^d \).

With each block $F_i$ being $L_i$-Lipschitz in the input that satisfies follows.:
\begin{equation}\label{append:L_Lip}
    ||F_i(z;\theta_i)-F_i(z';\theta_i)|| \le L_i||z-z'||
\end{equation}
And $M_i$-Lipschitz in the parameters which satisfies follows.:
\begin{equation}\label{append:M_Lip}
    ||F_i(z;\theta_i)-F_i(z;\theta_i')|| \le M_i||\theta_i-\theta_i'||
\end{equation}
Then, for any two parameter sets \( \theta = (\theta_1, \dots, \theta_L) \) and \( \hat{\theta} = (\hat{\theta}_1, \dots, \hat{\theta}_L) \), the following holds:
\begin{equation}
||\mathcal{M}(x;\theta) - \mathcal{M}(x;\hat{\theta})|| \le \sum\limits^{L}_{i=1}\Bigg[\Bigg( \prod\limits^{L}_{k=i+1}(1+L_k)\Bigg)M_i ||\theta_i-\hat{\theta}_i||\Bigg] \end{equation}
\end{lemma}
\begin{proof}
With Eq.~\ref{append:proof_transformer}, we can formulate the difference of two hidden states between dense model and modified model as follows.:
\begin{equation}
    ||z_{i+1} - \hat{z}_{i+1}|| = ||[ z_i + F(z_i;\theta_i) ] - [ \hat{z}_i + F(\hat{z}_i;\hat{\theta}_i) ]||
\end{equation}
By the triangle inequality:
\begin{equation}
    ||z_{i+1} - \hat{z}_{i+1}|| \le ||z_i - \hat{z}_i|| + \underbrace{||F(z_i;\theta_i) - F(\hat{z}_i;\hat{\theta}_i)||}_{\text{(A)}}
\end{equation}
We now split term (A) with the assumptions in Eq.~\ref{append:L_Lip} and Eq.~\ref{append:M_Lip}:
\begin{align}
    \text{(A)} &= ||F(z_i;\theta_i) - F(\hat{z}_i;\hat{\theta}_i)|| \\
               &\le ||F_i(z_i;\theta_i)-F_i(\hat{z};\theta_i)|| + ||F_i(\hat{z}_i;\theta_i)-F_i(\hat{z}_i;\hat{\theta}_i)|| \\
               &\le \underbrace{L_i||z_i - \hat{z}_i||}_{\text{input Lipschitz}} + \underbrace{M_i||\theta_i - \hat{\theta}_i||}_{\text{parameter Lipschitz}}
\end{align}
So combining,
\begin{equation}
    ||z_{i+1} - \hat{z}_{i+1}|| \le ||z_i - \hat{z}_i|| + L_i||z_i - \hat{z}_i|| + M_i||\theta_i-\hat{\theta}_i||
\end{equation}
Hence, 
\begin{equation}\label{append:inter_ineq}
    ||z_{i+1} - \hat{z}_{i+1}|| \le (1+L_i)||z_i - \hat{z}_i|| + M_i||\theta_i-\hat{\theta}_i||
\end{equation}
Define the error at block $i$ as
\begin{equation}
    E_i=||z_i - \hat{z}_i||
\end{equation}
Eq.~\ref{append:inter_ineq} becomes
\begin{equation}
    E_{i+1} \le (1+L_i)E_i + M_i||\theta_i-\hat{\theta}_i||
\end{equation}
We start from $E_1=||z_1 - \hat{z}_1||=||x-x||=0$ (since both netowrks see the same input $x$). Thus:
\begin{align}
    E_2 &\le (1+L_1)E_1 + M_1 ||\theta_1 - \hat{\theta}_1|| = M_1 ||\theta_1 - \hat{\theta}_1|| \\
    E_3 &\le (1+L_2)E_2 + M_2 ||\theta_1 - \hat{\theta}_1|| \le (1+L_2)[M_1||\theta_1 - \hat{\theta}_1|| + M_2||\theta_2 - \hat{\theta}_2||]
\end{align}
If we do recursive telescoping over all blocks, we get
\begin{equation}
    E_{L+1} = ||z_{L+1} - \hat{z}_{L+1}|| \le \sum\limits^{L}_{i=1}\Bigg[\Bigg( \prod\limits^{L}_{k=i+1}(1+L_k)\Bigg)M_i ||\theta_i-\hat{\theta}_i||\Bigg]
\end{equation}
Since $\mathcal{M}(x;\theta) = z_{L+1}$ and $\mathcal{M}(x;\hat{\theta})=\hat{z}_{L+1}$, we have shown:
\begin{equation}
    ||\mathcal{M}(x;\theta) -\mathcal{M}(x;\hat{\theta})|| = ||z_{L+1} - \hat{z}_{L+1}|| \le \sum\limits^{L}_{i=1}\Bigg[\Bigg( \prod\limits^{L}_{k=i+1}(1+L_k)\Bigg)M_i ||\theta_i-\hat{\theta}_i||\Bigg]
\end{equation}

\end{proof}
\subsection{Proof of Theorem~\ref{thm:feasibility}}\label{append:proof_thm}
\renewcommand{\thetheorem}{3.2}
\begin{theorem}[Tighter error bound of Re-use] Under the same assumptions in Lemma~\ref{lemma:error_bound}, let $\theta_i$ denote the $i$-th block of a transformer that is skipped, $\theta^*_i$ represent the corresponding Re-used block, $U_{\text{Skip}}$ is a error bound for compressed model with Skip alone, and $U_{\text{Skip, Re-use}}$ is a error bound for a compressed model with skip and reuse. If following condition is satisfied,
    \begin{equation}\label{append:condition}
        ||\theta_i - \theta^*_i|| < ||\theta_i||
    \end{equation}
then, the following inequality holds: \begin{equation}
        U_{\text{Skip, Re-use}} < U_{\text{Skip}}
    \end{equation}
\end{theorem}
\begin{proof}
With Lemma~\ref{lemma:error_bound}, we can formulate the error bound of compressed model as follows:
\begin{equation}
    ||\mathcal{M}(x;\theta) - \mathcal{M}(x;\hat{\theta})|| \le \sum\limits^{L}_{i=1}C_i||\theta_i-\hat{\theta}_i||
\end{equation}
where $C_i$ is a constant for each $i$-th block. For the unskipped block, $\theta_i=\hat{\theta}_i$, $||\theta_i-\hat{\theta}_i||=0$. So, we define a parameter set $\hat{\theta}_{\text{Skip}}$ that exclude $\theta_i$ in the parameter set $\theta$. And we can rewrite the Lemma~\ref{lemma:error_bound} as follows:
\begin{equation}\label{append:lemma_inter_err}
    ||\mathcal{M}(x;\theta) - \mathcal{M}(x;\hat{\theta}_{\text{Skip}})|| \le \sum_{i\in \mathcal{S}}C_i||\theta_i-\hat{\theta}_i||
\end{equation}
where $\mathcal{S}$ is a set of skipped (pruned) block indices. And skipped block can be represented as follows:
\begin{equation}
    \hat{\theta}_{i,\text{Skip}}=\mathbf{0}
\end{equation}
Then, we can manipulate the error bound Eq.~\ref{append:lemma_inter_err} with following.
\begin{equation}
    ||\mathcal{M}(x;\theta) - \mathcal{M}(x;\hat{\theta}_{\text{Skip}})|| \le \sum_{i\in \mathcal{S}}C_i||\theta_i|| = U_{\text{Skip}}
\end{equation}
Re-use substitute the skipped weight to the weight of adjacent block:
\begin{equation}
    \hat{\theta}_{i,\text{Re-use}}=\theta^*_i
\end{equation}
If we make set of $i$ that satisfies Eq.~\ref{append:condition} and denote as $\mathcal{R}$ then apply Re-use,
\begin{equation}
    ||\mathcal{M}(x;\theta) - \mathcal{M}(x;\hat{\theta}_{\text{Skip, Re-use}})|| < \sum_{i\in \mathcal{S} \land i\notin \mathcal{R}}C_i||\theta_i|| + \sum_{j\in \mathcal{R}}C_j||\theta_j - \theta^*_j|| = U_{\text{Skip, Re-use}}
\end{equation}
Since the $\mathcal{R}$ consists of block indices that satisfies Eq.~\ref{append:condition}, we have shown:
\begin{equation}
    U_{\text{Skip, Re-use}} = \sum_{i\in \mathcal{S} \land i\notin \mathcal{R}}C_i||\theta_i|| + \sum_{j\in \mathcal{R}}C_j||\theta_j - \theta^*_j|| < \sum_{k\in \mathcal{S}}C_k||\theta_k|| = U_{\text{Skip}}
\end{equation}
\end{proof}

\section{Detailed Experimental Setup}
\subsection{Baseline configurations}
\label{append:baselines}
\begin{table}[H]
    \caption{Block indices ordered by the Block Influence (BI) score obtained from ShortGPT with our calibration dataset. Blocks with lower BI scores are ranked earlier, while those with higher BI scores are ranked later. During block pruning, blocks with the lowest BI scores are pruned first, according to the specified pruning ratio.}
    \label{append_tab:shortgpt}
    \centering
    \begin{tabular}{c}
    \toprule
    Order of block index  \\
    \midrule
    10, 11, 8, 9, 12, 13, 5, 14, 6, 4, 3, 7, 15, 17, 16, 18, 2, 19, 20, 21, 22, 1, 23, 0 \\
    \bottomrule
    \end{tabular}
\end{table}
\paragraph{ShortGPT~\cite{men2024shortgpt}}
ShortGPT calculates the Block Influence (BI) score by measuring cosine similarity between intermediate features of each layer, extracted from the calibration data, and then taking its complement. We calculated the BI scores using the 1k calibration subset of the CC12M dataset that we constructed and sorted the blocks in ascending order based on their BI scores. The ordered block indices are presented in Table~\ref{append_tab:shortgpt}.

\paragraph{LaCo~\cite{yang2024laco}}
LaCo presents an algorithm that merges adjacent transformer layers on the basis of the cosine similarity between their output features and those of the original model. If similarity exceeds a predefined threshold, the layers are merged to reduce the size of the model. However, our experiments revealed that LaCo's performance is highly sensitive to its hyperparameter settings. In some cases, the algorithm failed to effectively reduce the number of parameters.

To ensure a fair comparison, we conducted thorough and extensive experiments to identify the optimal hyperparameters that maximize LaCo's performance in the Text-to-Image (T2I) diffusion model. The selected hyperparameters are as follows:
Layer collapse interval $\mathcal{I}=2$, Number of layers to merge $\mathcal{M}=2$, First layer to merge $\mathcal{L}=1$, Last layer to merge $\mathcal{H}=24$, cosine similarity threshold $\mathcal{T}=0.7$.

We applied LaCo by continuously merging layers until the compressed model achieved the desired target sparsity.

\paragraph{FinerCut~\cite{zhang2024finercut}}
\begin{table}[!h]
    \caption{Sub-block indices ordered with MSE metrics from FinerCut with our calibration dataset. For the efficiency, we extracted top-24 sub-block indices for pruning. During the pruning process, blocks with the lowest MSE are pruned first, according to the specified pruning ratio.}
    \label{append_tab:finercut}
    \centering
    \begin{tabular}{c}
    \toprule
    Order of sub-block index  \\
    \midrule
    46, 5, 11, 6, 47, 44, 20, 19, 18, 45, 40, 41, 1, 32, 31, 17, 24, 23, 3, 42, 43, 30, 38, 37\\
    \bottomrule
    \end{tabular}
\end{table}
FinerCut employs a finer-grained blockwise pruning strategy based on the structural decomposition of transformer blocks into two distinct sub-blocks: Multi-Head Attention (MHA) and Feed-Forward Network (FFN) sub-blocks. To assess the importance of each sub-block, FinerCut originally proposed various evaluation metrics: 1) cosine similarity, 2) mean-squared error (MSE), and 3) Jensen-Shannon divergence (JSD). In its original implementation for auto-regressive LLMs, FinerCut adopted JSD after evaluating the effectiveness of these metrics. However, since the text encoder in the T2I diffusion model lacks a language modeling head, metrics based on perplexity and JSD could not be applied. For the fair comparison, we implemented FinerCut using MSE as the sole metric to evaluate sub-block importance, which is the closest metric with our discrepancy metric. We also conducted experiments of FinerCut with cosine similarity in Sec.~\ref{append:another_metric}.

The sorted sub-block indices determined by FinerCut are presented in Table~\ref{append_tab:finercut}. The even indices represents the MHA blocks and the odd indices denote the FFN blocks. Due to FinerCut’s sub-block-level pruning granularity, it was not possible to achieve the exact sparsity ratios as ShortGPT and LaCo. Therefore, pruning was carried out up to the sub-block index that most closely matched the target sparsity for a fair comparison.

\paragraph{Skrr (Ours)}
This section details the order of indices determined during the Skip phase of Skrr. Because the projection module is incorporated into the denoising module of each diffusion model, the sub-block indices vary across different diffusion models. We executed the Skip phase for all models, and the resulting order of indices for each model is presented in the Table~\ref{append_tab:skrr_index_order}. This provides insight into how the indices are prioritized during the Skip phase for various diffusion architectures.

Additionally, the re-use indices of Skrr in PixArt-$\Sigma$ are presented in Table~\ref{append_tab:reuse_skrr_1}, Table~\ref{append_tab:reuse_skrr_2}, and Table~\ref{append_tab:reuse_skrr_3}. The re-use indices were independently computed using the skip indices for each sparsity level, and overall, the later blocks exhibited a tendency not to be reused.
\begin{table}[!h]
    \caption{Sub-block indices ordered with discrepancy metric $D$ from Skrr with our calibration dataset in PixArt-$\Sigma$, Stable Diffusion 3 (SD3), and FLUX.1-dev. For the efficiency, we extracted top-24 sub-block indices for pruning. During the pruning process, blocks with the lowest discrepancy are pruned first, according to the specified pruning ratio.}
    \label{append_tab:skrr_index_order}
    \centering
    \begin{tabular}{c|c}
    \toprule
    Model & Order of sub-blocks \\
    \midrule
    PixArt-$\Sigma$ & 46, 5, 6, 47, 44, 2, 11, 24, 23, 17, 18, 45, 10, 30, 29, 40, 39, 38, 37, 42, 25, 22, 43, 36 \\
    SD3 & 46, 5, 45, 6, 11, 2, 22, 23, 17, 18, 30, 29, 43, 44, 20, 1, 41, 42, 40, 39, 38, 37, 21, 16 \\
    FLUX.1-dev &  46, 45, 5, 6, 11, 20, 19, 22, 23, 30, 29, 47, 2, 44, 32, 31, 18, 17, 37, 36, 43, 40, 39, 42 \\
    \bottomrule
    \end{tabular}
\end{table}

\begin{table}[h]
    \centering
    \begin{minipage}{0.3\textwidth}
        \centering
        \begin{tabular}{c|c}
        \toprule
            \textbf{Skipped Block} & \textbf{Re-used Block} \\ \hline
            2  & 4 \\
            5  & 7 \\
            6  & - \\
            11 & 13 \\
            17 & 19 \\
            18 & 16 \\
            23 & 21 \\
            24 & - \\
            44 & - \\
            45 & - \\
            46 & - \\
            47 & - \\
        \bottomrule
        \end{tabular}
        \caption{Skrr Re-use indices of sparsity level 1 in PixArt-$\Sigma$.}
        \vspace{3.8cm}
        \label{append_tab:reuse_skrr_1}
    \end{minipage}
    \begin{minipage}{0.3\textwidth}
        \centering
        \begin{tabular}{c|c}
            \toprule
            \textbf{Skipped Block} & \textbf{Re-used Block} \\ \hline
            2  & 4 \\
            5  & 7 \\
            6  & 4 \\
            10  & - \\
            11  & 13 \\
            17  & 19 \\
            18 & 20 \\
            23 & 21 \\
            24 & - \\
            29  & - \\
            30  & - \\
            39  & - \\
            40  & - \\
            44 & - \\
            45 & - \\
            46 & - \\
            47 & - \\
            \bottomrule
        \end{tabular}
        \caption{Skrr Re-use indices of sparsity level 2 in PixArt-$\Sigma$.}
        
        \vspace{1.7cm}\label{append_tab:reuse_skrr_2}
    \end{minipage}
    \begin{minipage}{0.3\textwidth}
        \centering
        \begin{tabular}{c|c}
        \toprule
            \textbf{Skipped Block} & \textbf{Re-used Block} \\ \hline
            2  & 0 \\
            5  & 7 \\
            6  & 4 \\
            10  & 12 \\
            11  & 9 \\
            17  & 19 \\
            18 & 16 \\
            23 & 21 \\
            24 & - \\
            25 & 27 \\
            29  & - \\
            30  & - \\
            37  & - \\
            38  & - \\
            39  & - \\
            40  & - \\
            42  & - \\
            44 & - \\
            45 & - \\
            46 & - \\
            47 & - \\
        \bottomrule
        \end{tabular}
        \caption{Skrr Re-use indices of sparsity level 3 in PixArt-$\Sigma$.}
        \label{append_tab:reuse_skrr_3}
    \end{minipage}
\end{table}

\subsection{Diffusion model and text encoder}
\label{append:diff_and_text}
For the quantitative comparison, we performed experiments on two diffusion transformer~\cite{peebles2023scalable} (DiT) text encoders: PixArt-$\Sigma$~\cite{chen2025pixart}. PixArt-$\Sigma$ employs the T5-XXL model~\cite{raffel2020exploring} as its text encoder. We have also conducted experiments on compressing text encoders that leverages several text encoders. For example, Stable Diffusion 3 (SD3) leverages CLIP-L, CLIP-G, and T5-XXL. The results of compressing multiple text encoders are presented in the Sec.~\ref{append:multiple_text_encoder}. For quantitative and qualitative evaluations, we measured discrepancy and generated images using Bfloat16 precision with the pretrained weights \texttt{PixArt-alpha/PixArt-Sigma-XL-2-1024-MS} obtained from the Hugging Face Diffusers library. For PixArt-$\Sigma$, all images were generated in the resolution of $512\times512$.  Furthermore, the FLUX model and SD3, included in the qualitative results, was evaluated using the \texttt{stabilityai/stable-diffusion-3-medium} and \texttt{black-forest-labs/FLUX.1-dev} weights from the Hugging Face Diffusers library. The text encoder configuration and inference precision for both models were consistent with the aforementioned setup, using the Bfloat16 precision and applying the same pruning strategy to ensure a fair and consistent evaluation. Additionally, we fixed the number of function evaluations (NFE) to 20 across all diffusion models. All other hyperparameters, such as the classifier-free guidance~\cite{ho2021classifier} scale, were set to the default value. All experiments were performed on a single Nvidia A100 or RTX 4090 GPU.

\subsection{Calibration dataset}\label{append:calibration}
We constructed a calibration set derived from the CC12M~\cite{changpinyo2021conceptual} dataset to identify blocks for pruning. While existing LLM-based methods typically utilize calibration sets such as the Common Crawl's web corpus~\cite{raffel2020exploring} (C4), SlimPajama~\cite{cerebras2023slimpajama} and WikiText~\cite{merity2016pointer}, primarily focusing on perplexity-based performance metrics, these approaches are not directly applicable to T2I models. To address this gap, we curated a calibration set specifically tailored for the T2I text encoder by selecting only clean captions and semantically rich prompts ranging from 150 to 250 tokens from the CC12M image-text paired dataset. This selection ensures a more effective calibration for image synthesis tasks. Representative examples of prompts from the constructed dataset are provided in Table~\ref{append_tab:calibration}.

\begin{table}[!h]
    \caption{Examples of prompts from the calibration set across various lengths are presented. These text prompts are sampled from the CC12M dataset, featuring rich and descriptive expressions with lengths optimized for calibration. This selection ensures the prompts are semantically meaningful and well-suited for effective text-to-image model calibration.}
    \label{append_tab:calibration}
    \centering
    \begin{tabular}{p{0.95\textwidth}}
    \toprule
    \multicolumn{1}{c}{Example prompts in calibration dataset} \\
    \midrule    
    1. A collection of photography equipment neatly arranged on a wooden surface. 
    Items include a camera, smartphone, tablet, drone, portable power bank, tripod, cleaning kit, strap, case, and backpack. 
    The warm wooden background contrasts with the modern gear. \\
    
    2. A moment at a subway station with a vintage train numbered ``7'' at the platform. 
    The platform has safety barriers and a yellow line, illuminated by fluorescent lights. 
    Text reads ``last stop for the 7 train'' and credits the photographer. \\
    
    3. A modern living room with a minimalist design. 
    A pendant light provides a warm glow. A wooden table holds a glass of water, a book, a smartphone, and a notebook. 
    A white cabinet and a cityscape view complete the cozy atmosphere. \\
    
    4. A person wearing a white t-shirt with the text ``A Day to Remember'' in pink and black lettering. 
    The shirt features a black collar and short sleeves, displayed plainly for product showcasing. \\
    
    5. A vibrant digital artwork of a stylized cityscape. 
    Buildings vary in color and pattern, resembling a patchwork quilt, creating a dense, lively urban environment. \\
    
    6. A small modern bathroom with brick-patterned walls and tiled flooring. 
    A white sink under a window, a glass shower enclosure, and a toilet create a rustic yet clean look. \\
    
    7. A vintage light-colored train car with blue and white stripes is parked on a track under a metal canopy. 
    Metal stairs lead to the entrance, possibly part of a museum exhibit. \\
    
    8. A wall with a playful quote: 
    ``In this house we are real, we make mistakes, we say I'm sorry, we give hugs, we give second chances, we forgive, we laugh a lot, we love each other, we are a family.'' 
    A guitar leaning against the wall adds a cozy, homey touch. \\
    
    9. A vibrant bouquet of flowers arranged in a clear glass vase. the bouquet consists of various types of flowers, including hydrangea, calla lilies, roses, and gerbera daisies, with burgundy berries interspersed among them. the flowers are in shades of pink and purple, creating a striking contrast. the arrangement is set against a plain, light-colored background, which accentuates the colors and textures of the flowers. the style of the image is a close-up photograph that captures the details of the floral arrangement. \\
    
    10. The memorial church at stanford university, a large, ornate building with a prominent cross at the top, illuminated at night. the facade is adorned with intricate mosaics and sculptures, including a central figure that appears to be a religious figure, possibly a saint or deity. the church's architecture is reminiscent of gothic and romanesque styles, with pointed arches and a large central archway that leads to the entrance. the surrounding area is dimly lit, with the church standing out as a beacon of light in the darkness. \\
    
    11. A serene lakeside setting with a houseboat that resembles a private yacht. the boat is equipped with a dining area featuring a table set for four with blue tableware, and a bar area with a blender, wine glasses, and a bottle of wine. the deck is furnished with multiple lounge chairs and a dining table, all under a retractable awning. the houseboat is docked near a rocky shoreline with a clear blue sky and a majestic red rock formation in the distance, suggesting a location like lake powell. the overall atmosphere is one of relaxation and leisure, ideal for a vacation or getaway. \\
    
    12. A graphic design with a stylized representation of a face, possibly a deity, with a serene expression. the face is framed by a green border with a white outline and a blue background. above the face, there is a crescent moon and a symbol that resembles a peace sign. below the face, the word ``chill'' is prominently displayed in bold, white capital letters. the overall style of the image is modern and graphic, with a clear emphasis on the word ``chill'' suggesting a theme of relaxation or tranquility.\\
    \bottomrule
    \end{tabular}
\end{table}

\subsection{Metrics}
\label{append:metrics}
\paragraph{Fréchet Inception Distance (FID)~\cite{heusel2017gans} score}
The Fréchet Inception Distance (FID) is a widely used metric for evaluating the performance of image generative models by quantifying the similarity between the distributions of real and generated images. Specifically, FID measures the Fréchet distance between feature representations extracted from a pre-trained image classification model, typically the Inception-V3 model. This approach leverages the model's rich intermediate features to capture high-level image statistics. The FID score is formally defined as follows:

\begin{equation}\label{append_eq:fid}
    d_F(\mathcal{N}(\mu, \Sigma), \mathcal{N}(\mu', \Sigma'))=||\mu - \mu'||^2_2 + \text{tr}{\Big(\Sigma + \Sigma' + 2(\Sigma\Sigma')^{\frac{1}{2}}\Big)}
\end{equation}
where $\mu$ and $\Sigma$ are the mean and covariance of the feature representations of real images, $\mu'$ and $\Sigma'$ are the mean and covariance of the feature representations of generated images. A lower FID score indicates that the generated images are more similar to the real images in both quality and diversity.

\paragraph{CLIP~\cite{radford2021learning} score}
The CLIP score evaluates the semantic alignment between a given text prompt and the image generated from that prompt by measuring the cosine similarity between their CLIP embeddings. This metric leverages a CLIP model trained on extensive image-text pairs to capture cross-modal relationships. The CLIP score can be formally defined as:
\begin{equation}\label{append_eq:clip}
    \text{CLIP}(I, T) = \cos{(E_{\text{image}}(I), E_{\text{text}}(T))} =  \frac{E_{\text{image}}(I) \cdot E_{\text{text}}(T)}{||E_{\text{image}}(I)|||E_{\text{text}}(T)||}
\end{equation}
where $I$ is the generated image, $T$ is the text prompt, $E_{\text{image}}(\cdot)$ represents the CLIP image encoder, and $E_{\text{text}}(\cdot)$ denotes the CLIP text encoder. The cosine similarity captures how well the generated image aligns with the semantic content of the text prompt. For our experiments, we leveraged the weights of the \texttt{openai/clip-vit-base-patch32} model from the Hugging Face library to calculate the CLIP score.

\paragraph{DreamSim~\cite{fu2023dreamsim} score}
The DreamSim score quantifies the semantic similarity at the mid-level between two images by evaluating their compositional, stylistic, and color characteristics. It quantifies how closely the overall structure and visual attributes of the images align. Formally, the DreamSim score can be expressed as Eq.~\ref{append_eq:dreamsim}:

\begin{equation}\label{append_eq:dreamsim}
    \text{DreamSim}(I_{\text{ref}}, I_{\text{pru}}) = 1 - \text{dist}_{\text{DreamSim}}(I_{\text{ref}}, I_{\text{pru}}), \quad \text{dist}_{\text{DreamSim}}(\cdot, \cdot)\in [0, 1]
\end{equation}
where $I_{\text{ref}}$ represents the image generated using the original text encoder, and $I_{\text{pru}}$ denotes the image generated using the pruned text encoder. The function $\text{dist}_{\text{DreamSim}}(\cdot, \cdot)$ computes the normalized semantic distance between the two images, as output by the DreamSim model. By subtracting this distance from 1, the DreamSim score reflects higher similarity with higher values, effectively capturing the semantic consistency between the original and pruned models' outputs.

\paragraph{GenEval~\cite{ghosh2024geneval}}
GenEval is a comprehensive evaluation metric designed to assess the degree to which a T2I generative model aligns generated images with input text prompts. In this study, GenEval was employed to evaluate whether the image synthesis results produced by the compressed text encoder accurately reflect the intended textual descriptions. The GenEval metric comprises six sub-metrics: \\
1. Single Object Generation – Assesses the model's ability to generate images from prompts containing a single object (e.g., \textit{``a photo of a giraffe''}). \\
2. Two Objects Generation – Evaluates the model's ability to correctly generate images from prompts with two distinct objects (e.g., \textit{``a photo of a knife and a stop sign''}). \\
3. Counting – Measures whether the model can accurately represent the specified number of objects (e.g. \textit{ ``a photo of three apples''}). \\
4. Colors - Verifies whether the generated image correctly reflects the color specified in the prompt (e.g., \textit{ ``a photo of a pink car''}). \\
5. Position – Tests the model's understanding of spatial relationships described in the prompt (e.g., \textit{``a photo of a sofa under a cup''}). \\
6. Color Attribution – Assesses the correct assignment of specified colors to multiple objects (e.g., \textit{``a photo of a black car and a green parking meter''}). \\
For evaluation, we generated images using a fixed random seed, producing 553 distinct prompts with four images per prompt, resulting in a total of 2,212 images. This setup ensured consistent and reproducible evaluation across all sub-metrics.

\subsection{Ablation study setup}
Due to constraints in our experimental environment, it was not feasible to evaluate the entire MS-COCO 30k image dataset for the ablation study. Instead, we performed the study using a 1k subset. Given that both the CLIP score and the DreamSim score effectively capture the impact of text encoder pruning, we focus our experiments on evaluating various configurations of these two metrics within the ablation study. Furthermore, for GenEval, experiments were conducted on the entire set. The experiments discussed in the Sec.~\ref{append:ablation} were conducted using the same experimental setup as described above.

\section{Additional Experiments}
\subsection{Interaction between blocks}
\begin{figure}[t]
  \centering
\includegraphics[width=\linewidth]{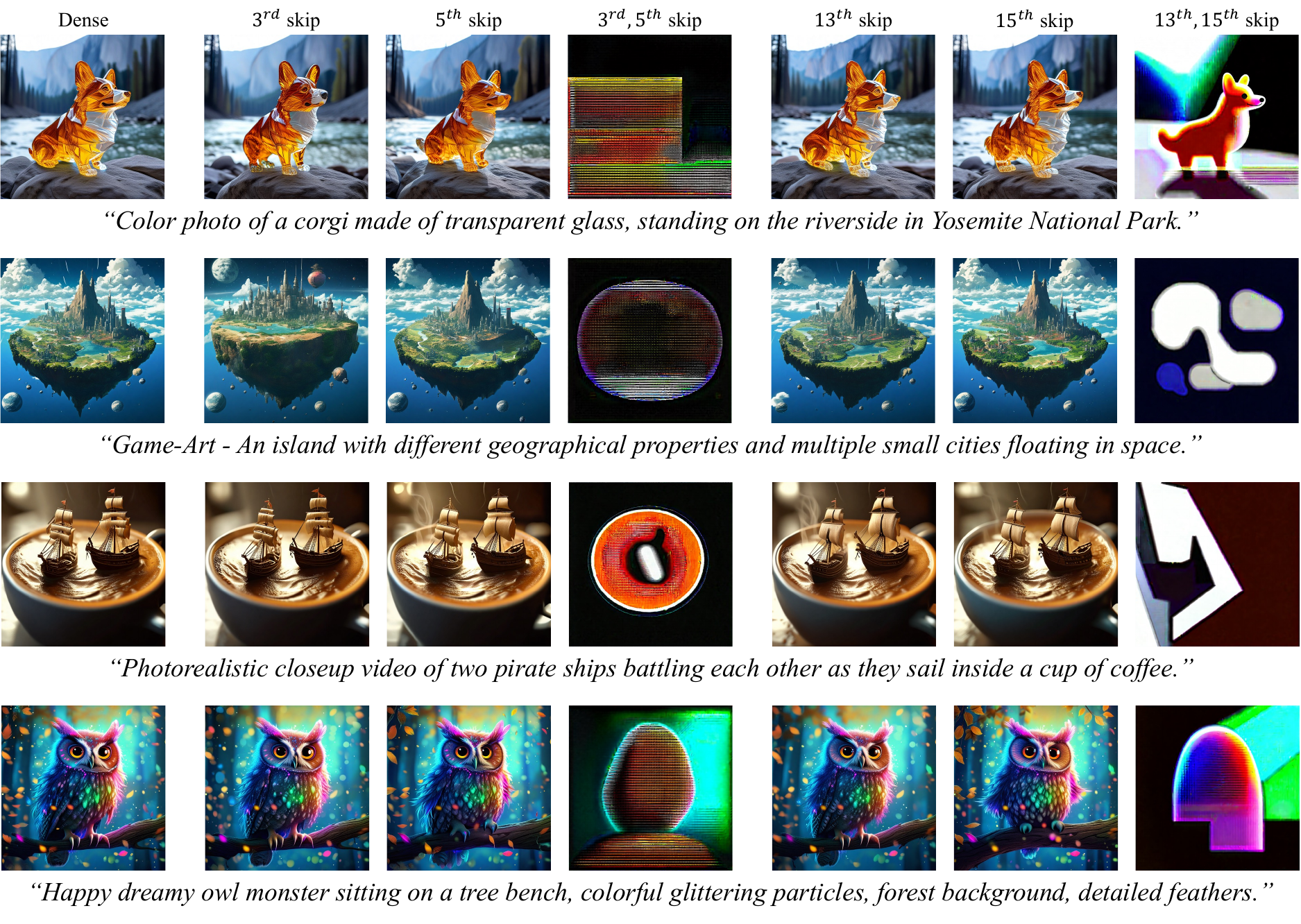}
\caption{
Examples illustrate various block interactions. When individual blocks are skipped, the generated images remain highly similar to those produced by the dense model, showing minimal impact. However, when two blocks are skipped simultaneously, the image is severely degraded, demonstrating the presence of significant interactions between blocks.
}
\label{append_fig:block_interaction}
\end{figure}

In addition to the block pairs shown in Fig.~\ref{fig:null_condition}, we identified additional pairs that exhibit interactions. These block pairs consistently maintain cosine similarity; however, their norms change significantly. We provide several qualitative results that highlight these interactions in Fig.~\ref{append_fig:block_interaction} which is experimented with PixArt-$\Sigma$. For cases where $k > 2$, the number of combinations increases to $\frac{48!}{k!(48-k)!}$, making exhaustive experimentation computationally infeasible. As a result, we limit our analysis to a subset of two-block interactions.

\begin{table}[t]
    \caption{Sparsity ratio of the text encoder, parameter count (Param.), memory usage (Mem.), FLOPs ratio of text encoder (T5-XXL) with respect to the total pipline, and total TFLOPs of the pipline were evaluated on SD3 across pruning methods, considering the entire pipeline to analyze each strategy's impact on computational cost. All metrics were measured at the maximum sparsity achieved by each pruning method and NFE with 28.}
    \label{append_tab:computation_sd3}
    \centering
    \begin{tabular}{cccccc}
    \toprule
    Method & Sparsity (\%) & Param. (B) & Mem. (GB) & FLOPs (\%) & TFLOPs \\
    \midrule
    Dense & 0.0 &7.66 & 14.49 & 0.41 & 174.2 \\
    \midrule
    ShortGPT & 40.5 & 5.73 & 10.93 & 0.32 & 174.1 \\
    LaCo & 40.5 & 5.73 & 10.82 & 0.32 & 174.1 \\
    FinerCut & 41.7 & 5.67 & 10.82 & 0.30 & 174.0 \\
    \textbf{Skrr (Ours)} & 41.9 & 5.73 & 10.93 & 0.35 & 174.1 \\
    \bottomrule
    \end{tabular}
\end{table}

\subsection{Computational cost on different diffusion models}
In the main paper, we report the computational cost only for PixArt-$\Sigma$. In this section, we extend the analysis to include the computational cost of dense, baselines, and Skrr-compressed models on Stable Diffusion 3 (SD3). Since SD3 contains more parameters and a more computationally intensive denoising module compared to PixArt-$\Sigma$, the text encoder consumes fewer FLOPs ratio in the total pipeline. However, the text encoder still accounts for significant memory usage. The detailed results are provided in Table~\ref{append_tab:computation_sd3}.
At the highest sparsity level, FinerCut achieves the lowest parameter count, memory usage, and FLOPs due to its slightly higher sparsity. Skrr exhibits the same parameter count and memory consumption as ShortGPT or LaCo at the same sparsity level (40.7\%). While Skrr incurs a slightly higher FLOP count than other baselines due to the re-use mechanism, the additional computational cost is minimal, accounting for less than 0.05\% of the entire pipeline or under 0.1 TFLOPS. Despite this, Skrr still demands less computation than the dense model.
These results underscore ability of Skrr to perform T2I synthesis in a computationally efficient manner, even when applied to models with varying complexities.

\subsection{Perturbation to the null condition feature}\label{append:perturb}
\begin{figure}[p]
  \centering
\includegraphics[width=1\linewidth]{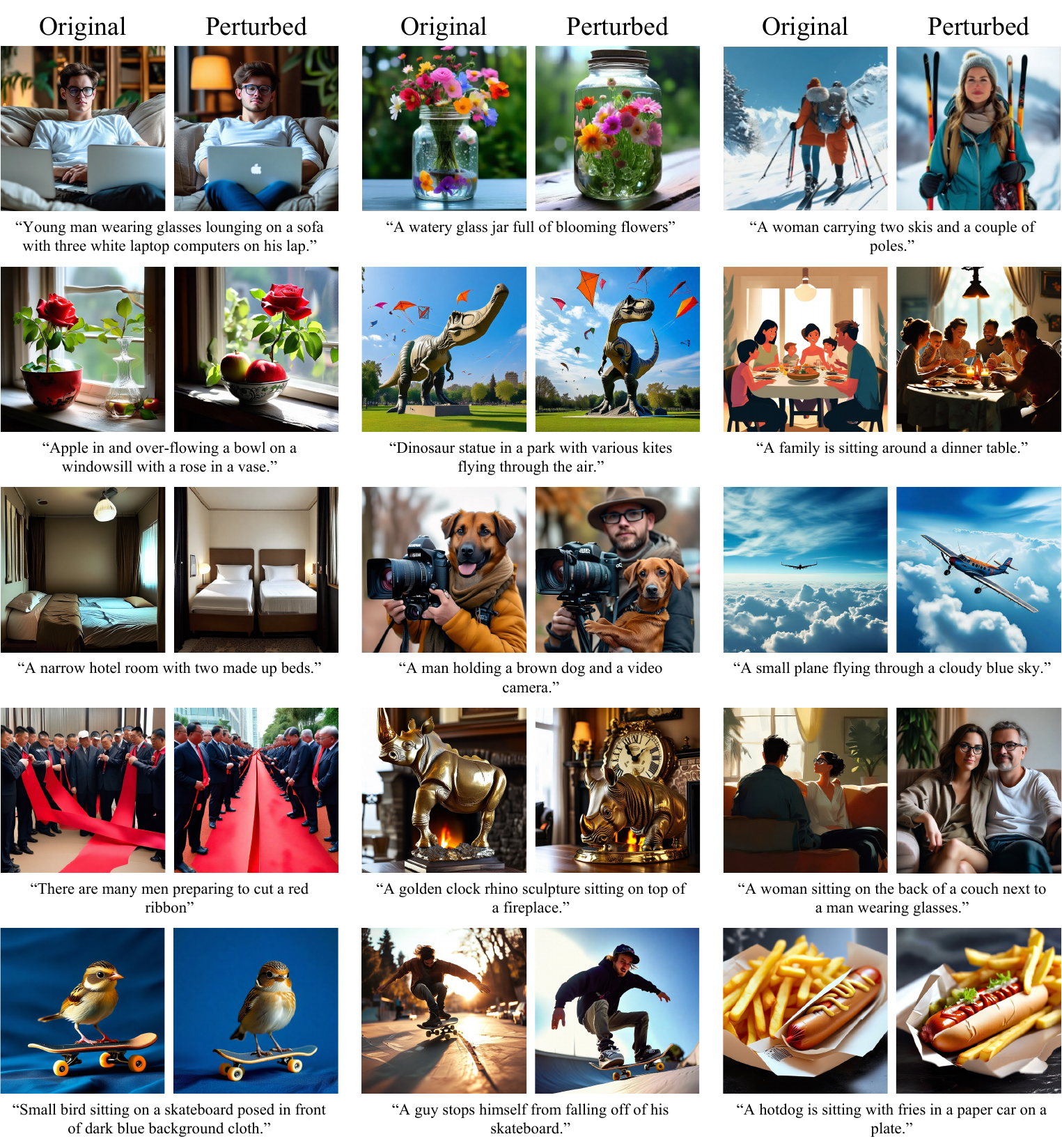}
\caption{Images generated from the same seed under the different null condition (original vs. perturbed) reveal notable differences. While the original model occasionally exhibits poor fidelity or omits objects specified in the prompt, the images generated using the perturbed null condition feature demonstrate higher fidelity, more accurate representation of the given prompt, and improved conditional image synthesis performance. All image pairs were generated with the same seed.}
\label{append_fig:perturb}
\end{figure}

We conducted an experiment to evaluate the impact of perturbations on $f_{\varnothing}$ and their effect on the FID and CLIP score. Using the PixArt-$\Sigma$ model, we applied a small scalar parameter $\lambda=10^{-2}$ and generated images for the MS-COCO 30k validation set. The results demonstrated that the perturbed $f_{\varnothing}$ produced a lower FID score ($22.89 \rightarrow 20.65$) and a comparable CLIP score ($0.314\rightarrow0.314$) to the original model. Furthermore, we provide a qualitative comparison between the images generated with the original null condition $f_{\varnothing}$ and those created using the perturbed feature vector $\hat{f}_{\varnothing}$ in Fig.~\ref{append_fig:perturb}.

\subsection{Applying Re-use to baselines}
\begin{figure}[p]
  \centering
\includegraphics[width=\linewidth]{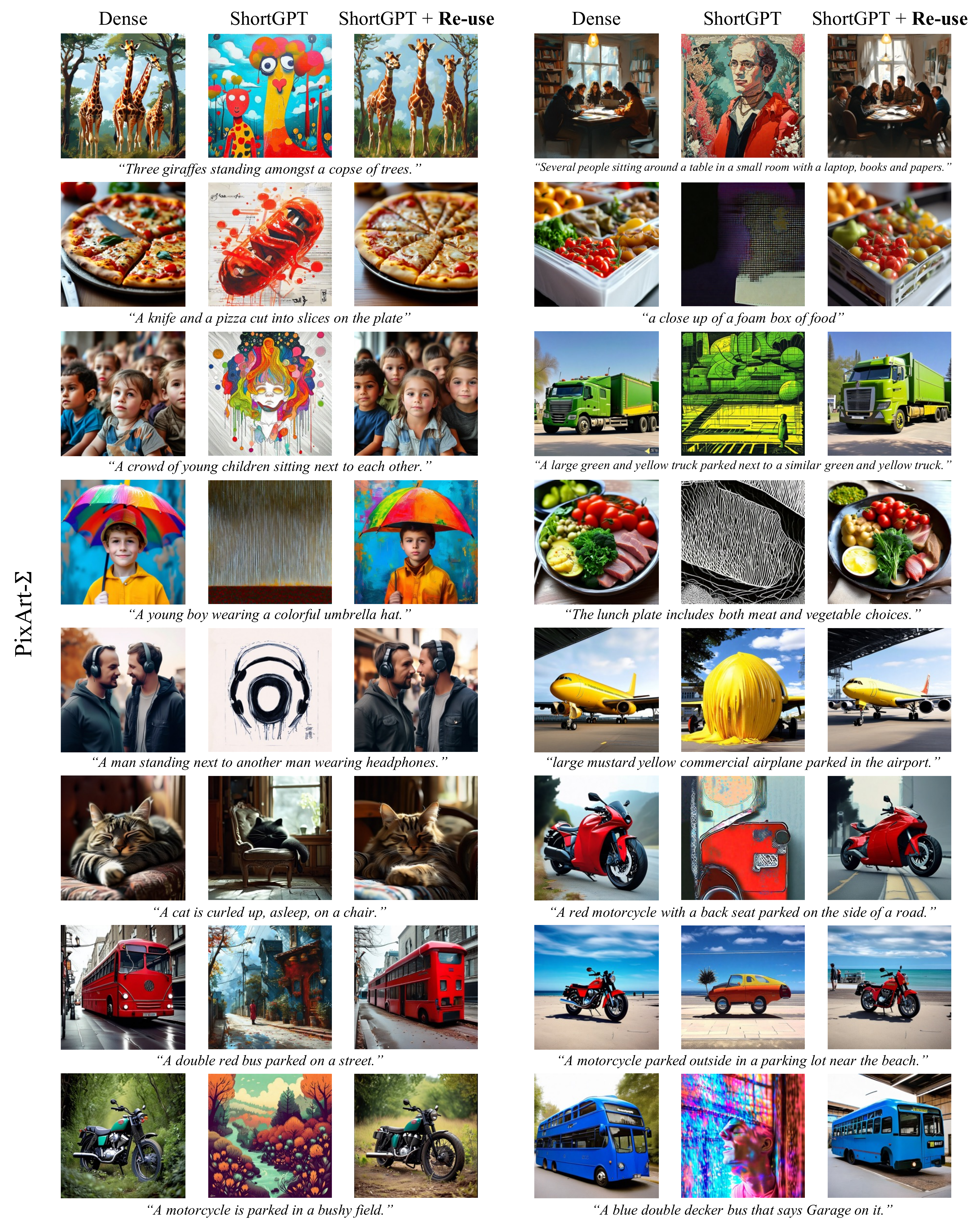}
\caption{Qualitative results of applying Re-use to ShortGPT. At high sparsity levels, ShortGPT struggles to generate high-fidelity images. Incorporating Re-use restores fidelity, text adherence, and alignment with the dense model.}
\label{append_fig:reuse_shorgpt}
\end{figure}

\begin{figure}[p]
  \centering
\includegraphics[width=\linewidth]{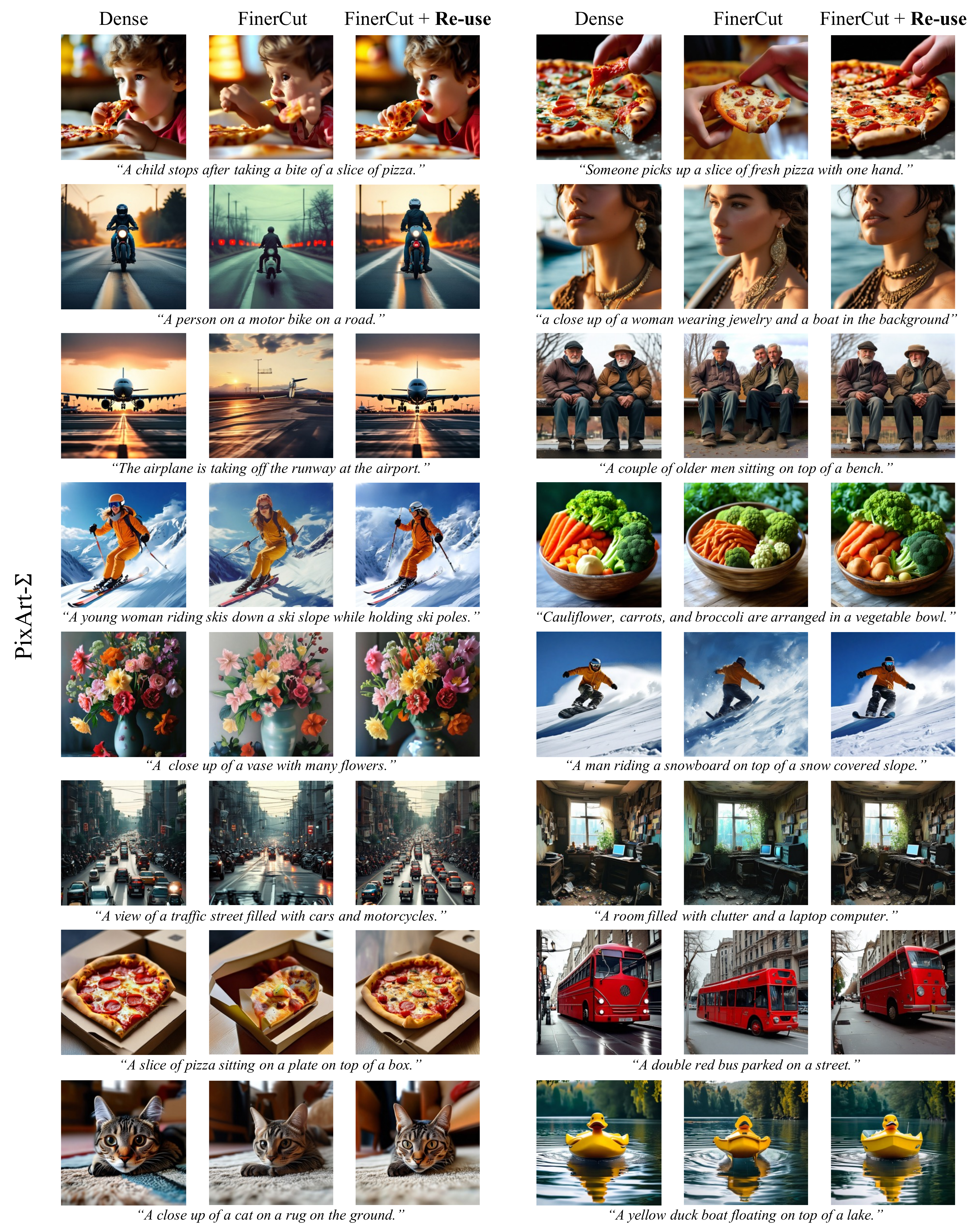}
\caption{
Qualitative results of applying Re-use to FinerCut. While FinerCut performs well compares other baselines in maintaining fidelity and text alignment, its alignment with the dense model decreases. Incorporating Re-use effectively restores this alignment.}
\label{append_fig:reuse_finercut}
\end{figure}

To further validate the effectiveness of Re-use, we conducted experiments to evaluate its compatibility with other block-wise pruning methods in a plug-and-play manner. Among the baselines considered for these experiments, ShortGPT and FinerCut employed block-pruning techniques, allowing us to apply Re-use and carry out the evaluations. ShortGPT prunes layer normalization, MHA, and FFN as a single unit, enabling us to perform Re-use on adjacent whole blocks. Similarly, FinerCut, which operates at the sub-block level like Skrr, also allowed the application of Re-use. The results of these experiments are presented in Fig.~\ref{append_fig:reuse_shorgpt} and Fig.~\ref{append_fig:reuse_finercut}. The blocks re-used by shortGPT and Finercut are presented in Table~\ref{append_tab:reuse_shortgpt} and Table~\ref{append_tab:reuse_finercut}, respectively. As shown in the results, Re-use demonstrated superior performance, further substantiating its effectiveness in alignment with both the experimental and theoretical findings presented earlier.

\begin{table}[h]
    \centering
    \begin{minipage}{0.3\textwidth}
        \centering
        \begin{tabular}{c|c}
        \toprule
            \textbf{Skipped Block} & \textbf{Re-used Block} \\ \hline
            4  & 7 \\
            5  & 7 \\
            6  & 7 \\
            8  & 7 \\
            9  & 7 \\
            10 & 7 \\
            11 & 7 \\
            12 & 7 \\
            13 & 7 \\
            14 & 7 \\
        \bottomrule
        \end{tabular}
        \caption{ShortGPT re-use indices}
        \vspace{4.2cm}
        \label{append_tab:reuse_shortgpt}
    \end{minipage}
    \hspace{2mm}
    \begin{minipage}{0.3\textwidth}
        \centering
        \begin{tabular}{c|c}
        \toprule
            \textbf{Skipped Block} & \textbf{Re-used Block} \\ \hline
            1  & 7  \\
            3  & -  \\
            5  & -  \\
            6  & 4  \\
            11 & 9  \\
            17 & 21 \\
            18 & 16 \\
            19 & 15 \\
            20 & 16 \\
            23 & 21 \\
            24 & -  \\
            31 & 29 \\
            32 & -  \\
            40 & -  \\
            41 & -  \\
            42 & -  \\
            44 & -  \\
            45 & -  \\
            46 & -  \\
            47 & -  \\
        \bottomrule
        \end{tabular}
        \caption{Finercut re-use indices}
        \label{append_tab:reuse_finercut}
    \end{minipage}
\end{table}

\subsection{Additional ablation study}\label{append:ablation}
\paragraph{Effectiveness of Re-use.}\label{append:re_use_quali}
In this section, we provide additional experimental results to demonstrate the effectiveness of Re-use. This mechanism effectively mitigates the performance degradation caused by Skip by utilizing the remaining blocks in the model. To evaluate its impact, we performed extensive quantitative and qualitative experiments. Quantitatively, we measured the CLIP, DreamSim, and GenEval scores of the PixArt-$\Sigma$ model with and without Re-use in the maximum sparsity. The results, presented in Table~\ref{append_tab:ablation_reuse}, show that Re-use enhances the CLIP, Dreamsim and GenEval scores. Although minor performance degradation was observed in other GenEval scores, the high performance achieved in challenging metrics like counting underscores the effectiveness of Re-use, especially in scenarios where the text encoder's capability plays a critical role.

\begin{figure}[p]
  \centering
\includegraphics[width=\linewidth]{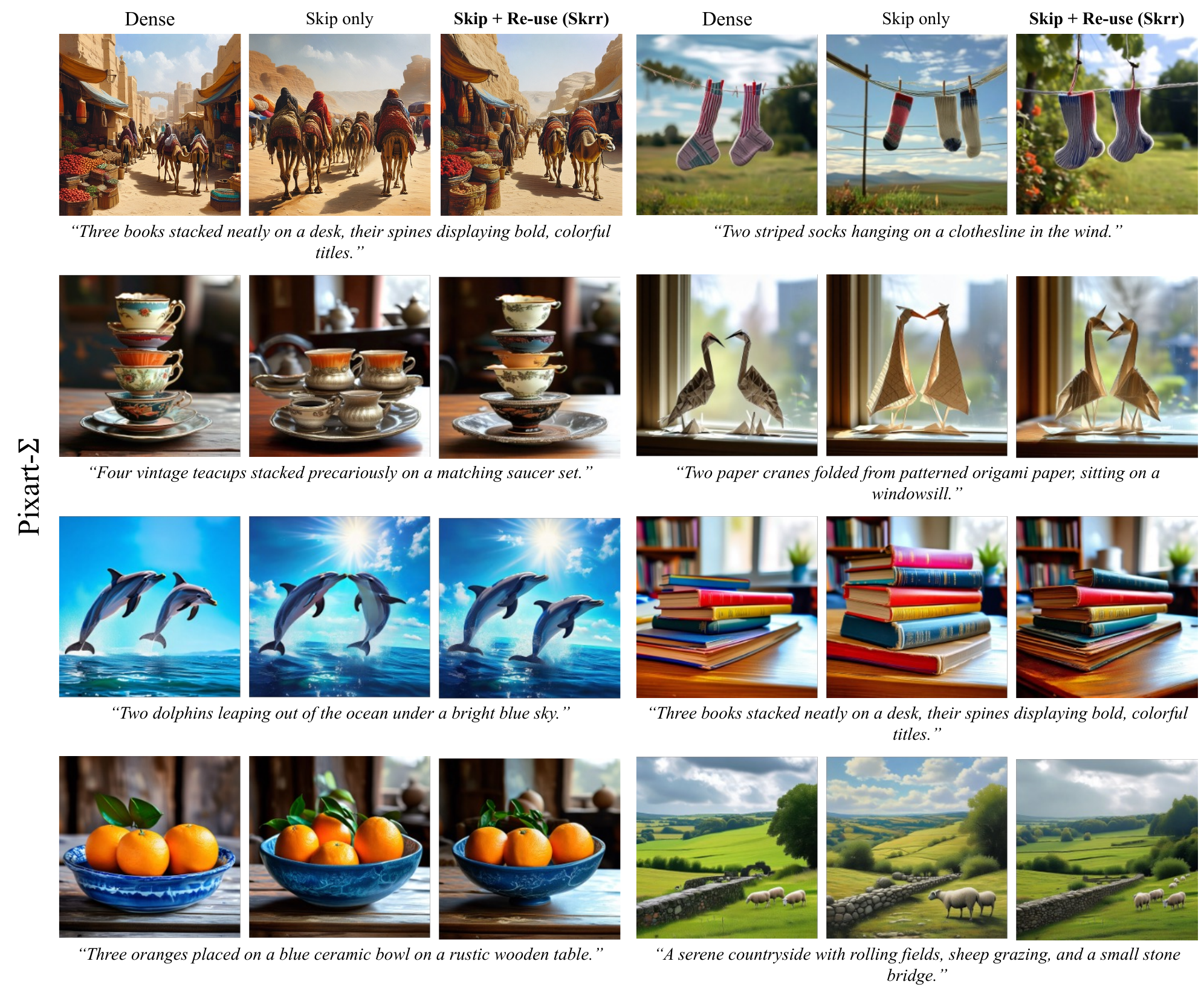}
\caption{Qualitative results on Re-use. Images generated with PixArt-$\Sigma$ and text encoder (T5-XXL) compressed by the full Skrr framework (dense, Skip only, and Skip with Re-use) are compared. With Skip only, performance noticeably degrades, particularly for detailed prompts or tasks requiring strong text encoder capabilities, such as counting, resulting in deviations from dense model outputs.}
\label{append_fig:reuse_pixart}
\end{figure}

\begin{figure}[p]
  \centering
\includegraphics[width=\linewidth]{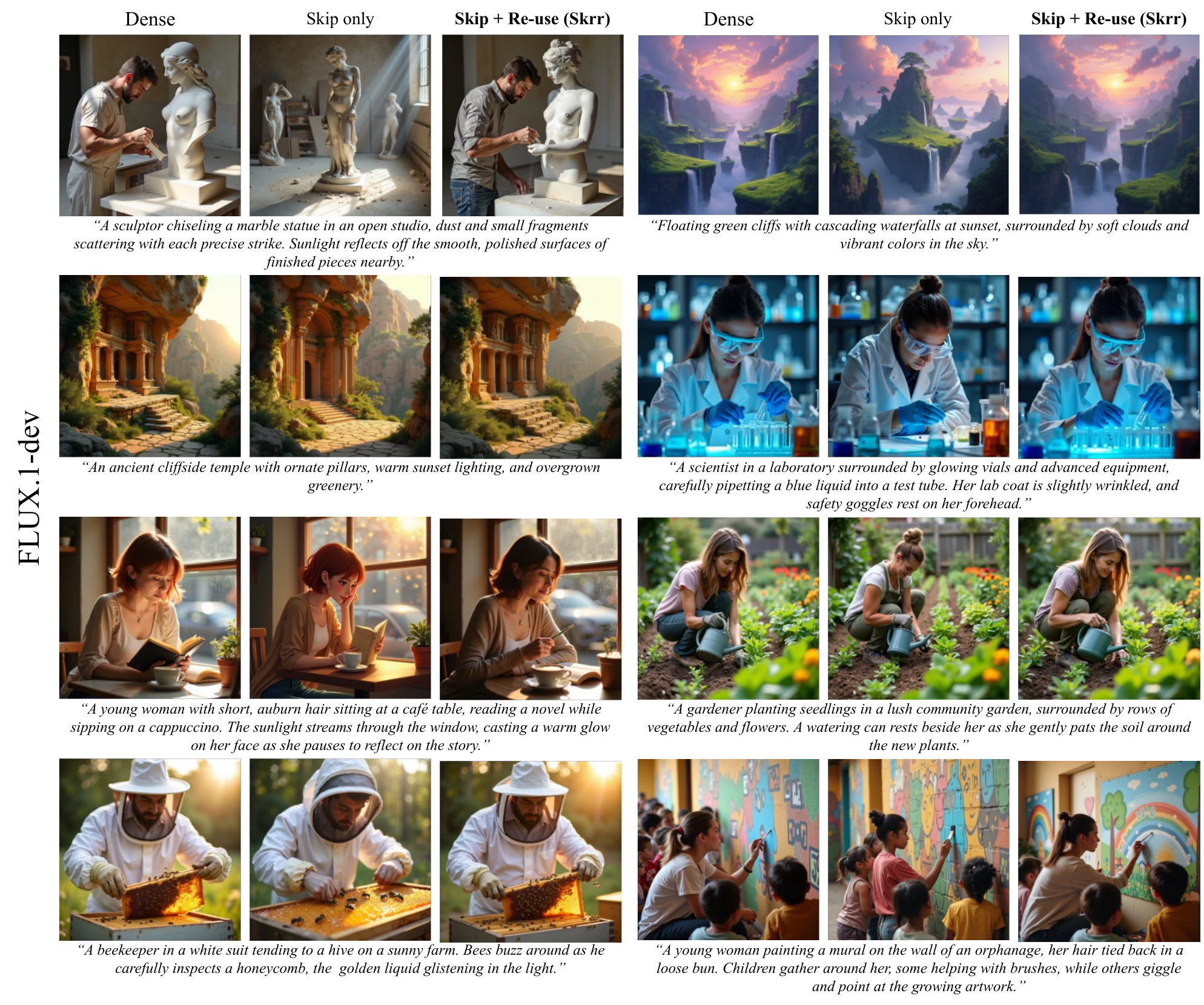}
\caption{Qualitative results on Re-use. Images generated with FLUX.1-dev and text encoder (T5-XXL) compressed by the full Skrr framework (dense, Skip only, and Skip with Re-use) are compared. When only Skip is used, some prompts may not be accurately reflected, leading to images that deviate from those generated by the dense model or even shift to an animated style rather than a realistic one.}
\label{append_fig:reuse_flux}
\end{figure}

In addition to the quantitative findings and qualitative examples presented in the main paper, we further confirmed that Re-use outperforms Skip alone in a variety of settings. These results are illustrated in Fig.~\ref{append_fig:reuse_pixart} for PixArt-$\Sigma$  and Fig.~\ref{append_fig:reuse_flux} for FLUX.1-dev. These visualizations reinforce the effectiveness of Re-use in generating images that align closely with text prompts, while maintaining model-agnostic behavior. The results demonstrate that Re-use not only enhances adherence to textual descriptions but also improves the performance of dense models across diverse scenarios.

\begin{table}[H]
    \caption{Quantitative ablation study for Re-use with PixArt-$\Sigma$. T2I synthesis performance with Re-use demonstrates that it effectively restores performance degraded by pruning, achieving excellent recovery without incurring additional memory overhead.}
    \label{append_tab:ablation_reuse}
    \centering
    \resizebox{0.8\linewidth}{!}{
    \begin{tabular}{cccccccccc}
    \toprule
    \multirow{2}{*}{Re-use} & \multirow{2}{*}{CLIP} & \multirow{2}{*}{DreamSim} & \multicolumn{7}{c}{GenEval} \\
    \cmidrule(lr){4-10}
    & & & Single & Two. & Count. & Colors & Pos. & Color attr. & Overall \\
    \midrule
    \xmark & 0.311  & 0.745 & 0.928 & 0.379 & 0.409 & 0.758 & 0.065 & 0.088 & 0.438 \\
    \cmark & \textbf{0.312}  & \textbf{0.746} & 0.913 & 0.409 & 0.450 & 0.755 & 0.055 & 0.068 & \textbf{0.442} \\
    \bottomrule
    \end{tabular}
    }
\end{table}

\paragraph{Compressing multiple text encoders.}\label{append:multiple_text_encoder}
Modern text-to-image (T2I) diffusion generative models frequently employ multiple text encoders to enhance their performance. A notable example is Stable Diffusion 3 (SD3), which incorporates CLIP-L, CLIP-G, and T5-XXL text encoders. In the experiments in the main article, we focused on compressing the T5-XXL text encoder, which is the largest and has the highest parameter count. Extending this approach, we evaluated the performance of the model when compressing all text encoders simultaneously. Specifically, for SD3, we applied compression to CLIP-L (30.6\% sparsity), CLIP-G (30.3\% sparsity) and T5-XXL (41.9\% sparsity).

\begin{figure}[p]
  \centering
\includegraphics[width=1\linewidth]{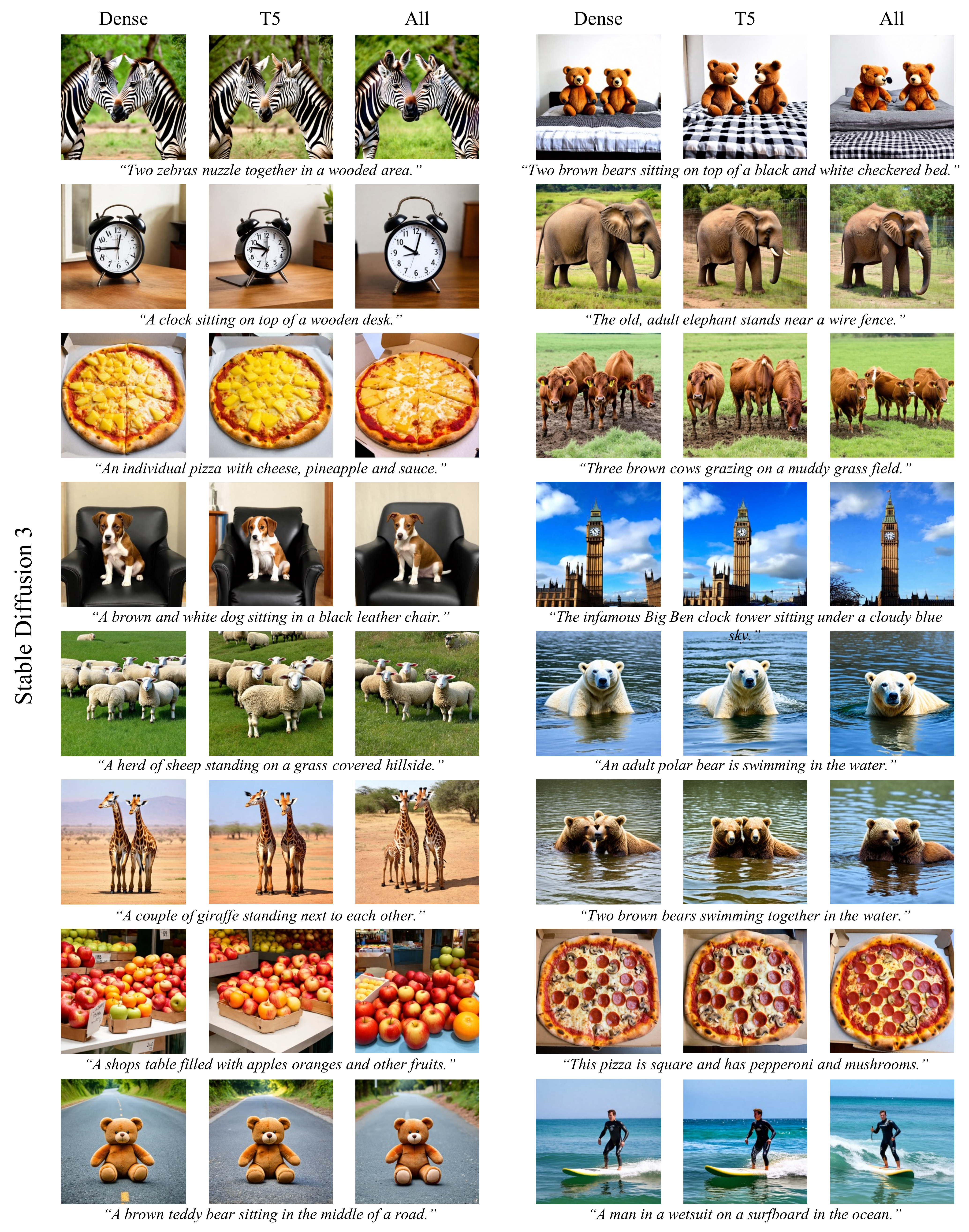}
\caption{Qualitative results of Skrr-compressed text encoders in Stable Diffusion 3. We compare compressing only the Dense model’s output and T5-XXL (T5) versus compressing all encoders (T5, CLIP-L, and CLIP-G). Skrr maintains high image fidelity and text-image alignment across both cases.}
\label{append_fig:multiple}
\end{figure}

The quantitative results, summarized in Table~\ref{append_tab:ablation_multiple}, reveal that although the CLIP and GenEval scores experience slight reductions, they remain sufficiently comparable to those of the dense model. Furthermore, qualitative results, illustrated in Fig.~\ref{append_fig:multiple}, demonstrate that images generated by the compressed model using Skrr on multiple text encoders exhibit remarkable similarity to those generated by the dense model. These findings indicate that compressing multiple text encoders simultaneously does not significantly compromise the model’s ability to generate high-quality, text-aligned images, thus validating the robustness of the compression approach.

\begin{table}[H]
    \caption{Quantitative ablation study for compressing multiple text encoders with Stable Diffusion 3. The second row presents the evaluation results of the model where only the T5 text encoder is compressed, while the third row corresponds to the model in which all three text encoders—T5, CLIP-L, and CLIP-G—are compressed.}
    \label{append_tab:ablation_multiple}
    \centering
    \resizebox{0.8\linewidth}{!}{
    \begin{tabular}{cccccccccc}
    \toprule
    \multirow{2}{*}{Compressed} & \multirow{2}{*}{CLIP} & \multirow{2}{*}{DreamSim} & \multicolumn{7}{c}{GenEval} \\
    \cmidrule(lr){4-10}
    & & & Single & Two. & Count. & Colors & Pos. & Color attr. & Overall \\
    \midrule
    Dense & 0.318  & 1.0 & 0.994 & 0.869 & 0.600 & 0.856 & 0.280 & 0.538 & 0.689 \\
    T5  & 0.317 & 0.811 & 0.959  & 0.773 & 0.500 & 0.803 & 0.210 & 0.423 & 0.611 \\
    All & 0.313  & 0.717 & 0.991 & 0.654 & 0.534 & 0.835 & 0.105 & 0.353 & 0.579 \\
    \bottomrule
    \end{tabular}
    }
\end{table}

\begin{figure}[p]
  \centering
\includegraphics[width=1\linewidth]{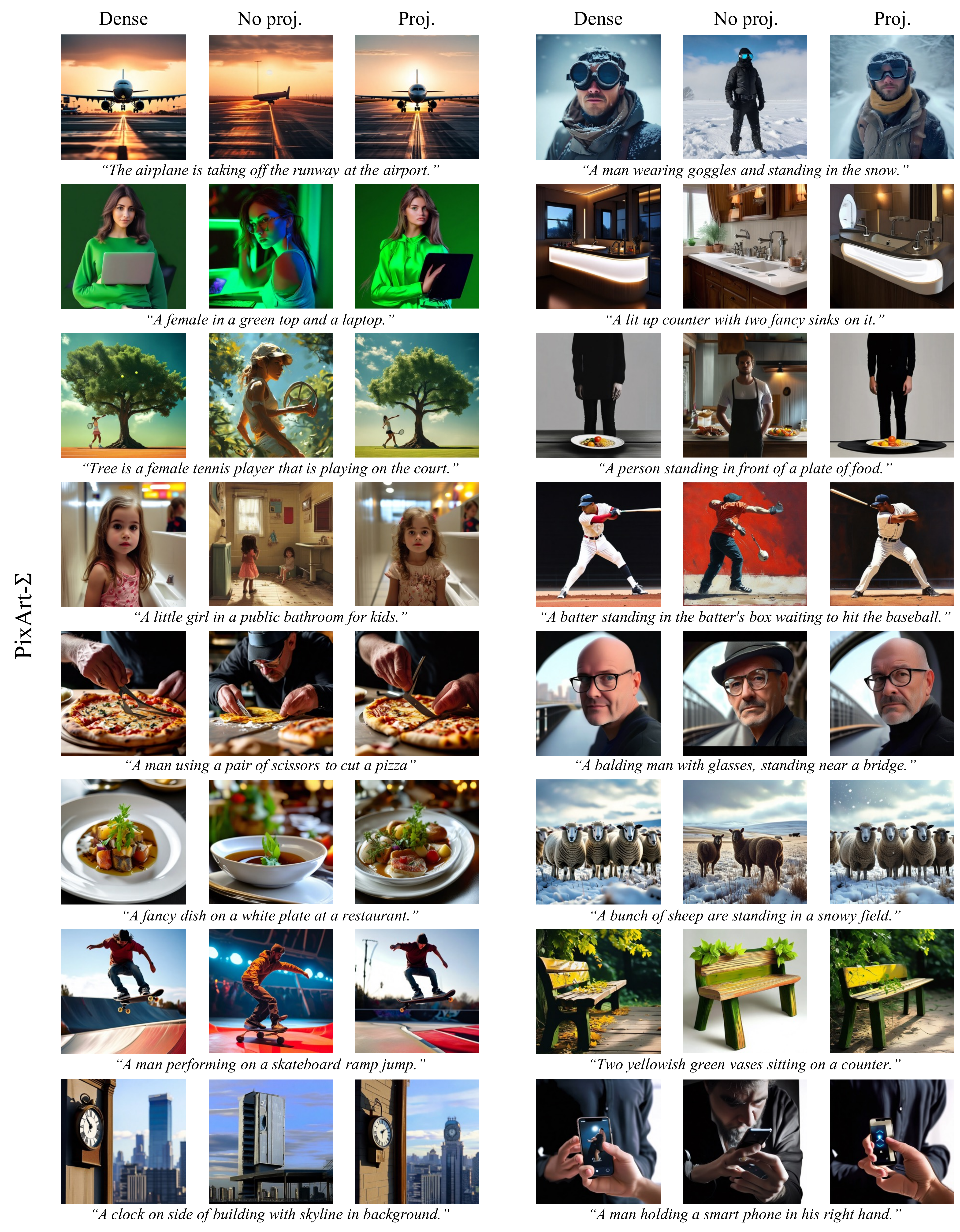}
\caption{Qualitative comparison of using a projection layer versus no projection in PixArt-$\Sigma$. Without projection (No Proj.), text-image alignment remains similar, but the generated image deviates from the dense model’s output. With projection (Proj.), both text-image alignment and image fidelity closely match the dense model.}
\label{append_fig:projection}
\end{figure}

\paragraph{Skrr without projection module.}
To measure the discrepancy in Skrr, we tailored the evaluation for the T2I diffusion model by analyzing features extracted from the text encoder after projection through the projection module used in the denoising process. In this section, we present both quantitative and qualitative results to validate the effectiveness of incorporating the projection module. We evaluated the performance of the PixArt-$\Sigma$ model compressed with Skrr, excluding the projection module. The results, summarized in Table~\ref{append_tab:ablation_proj}, reveal that, while performance is largely preserved without the projection module, a slight degradation occurs due to loss of information. This degradation occurs because the omission of the projection module ignores crucial components that are influential in the image generation process.

\begin{table}[H]
    \caption{Quantitative results of the ablation study on the projection module. Applying the projection improves CLIP score, DreamSim, and various GenEval tasks, demonstrating its significant impact on T2I generation performance when extracting features for discrepancy measurement.}
    \label{append_tab:ablation_proj}
    \centering
    \resizebox{0.8\linewidth}{!}{
    \begin{tabular}{cccccccccc}
    \toprule
    \multirow{2}{*}{Projection} & \multirow{2}{*}{CLIP} & \multirow{2}{*}{DreamSim} & \multicolumn{7}{c}{GenEval} \\
    \cmidrule(lr){4-10}
    & & & Single & Two. & Count. & Colors & Pos. & Color attr. & Overall \\
    \midrule
    \xmark & 0.305  & 0.708 & 0.884 & 0.366 & 0.266 & 0.620 & 0.070 & 0.078 & 0.381 \\
    \cmark & \textbf{0.312}  & \textbf{0.746} & 0.913 & 0.409 & 0.450 & 0.755 & 0.055 & 0.068 & \textbf{0.442} \\
    \bottomrule
    \end{tabular}
    }
\end{table}

This trend is also reflected in the qualitative results shown in Fig.~\ref{append_fig:projection}. The images generated with the projection module exhibit a higher degree of similarity to those produced by the dense model compared to the images generated without it. These findings underscore the importance of the projection module in preserving key features necessary for accurate text-to-image generation, thereby proving its effectiveness in maintaining model performance and fidelity.

\begin{figure}[p]
  \centering
\includegraphics[width=0.9\linewidth]{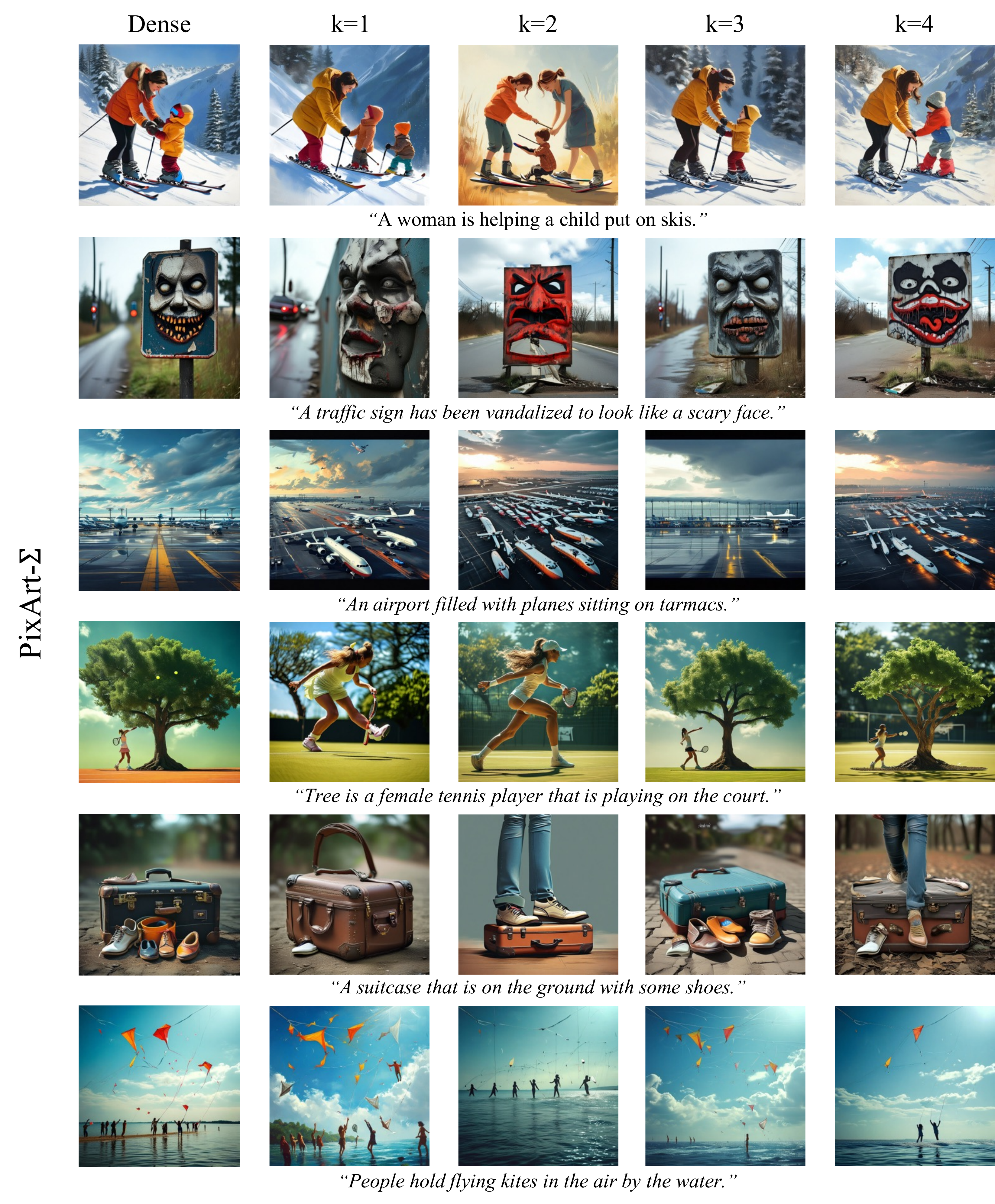}
\caption{Qualitative analysis of the impact of beam size $k$ on image generation quality. As shown, increasing the beam size enhances the alignment between generated images and the dense model, resulting in improved visual coherence and fidelity. Larger beam sizes allow for a more exhaustive exploration of the search space, leading to more refined and higher-quality outputs. These observations further support our quantitative findings presented in the main paper, demonstrating the effectiveness of using larger beam sizes.}
\label{append_fig:beamsearch}
\end{figure}

\paragraph{Qualitative results of beam size}
In the main manuscript, we quantitatively analyzed the performance variation with respect to the beam size $k$. Here, we provide qualitative results to further illustrate its impact on image generation. As shown in Fig.~\ref{append_fig:beamsearch}, increasing the beam size leads to better alignment between the generated images and the dense model, demonstrating the effectiveness of proper beam sizes in improving generation quality, but slight decline in performance as the beam size grows.

\subsection{Additional quantitative results}\label{append:another_metric}
FinerCut employs cosine similarity and Jensen-Shannon Divergence (JSD) in addition to the MSE used in our implementation. However, since the text encoder lacks a language head, JSD is not applicable. Moreover, FinerCut's reported results indicate that MSE outperforms cosine similarity, leading us to adopt it in our implementation. To ensure a fair comparison, we reimplemented FinerCut using cosine similarity and evaluated its performance using FID, CLIP, DreamSim, and GenEval, as presented in Table~\ref{append_tab:finercut_cosine}. Consistent with FinerCut’s findings, cosine similarity yielded better performance compared to MSE. Consequently, we used the MSE-based FinerCut implementation as a baseline for an equitable comparison that shows higher performance.

\begin{table*}[ht]
    \caption{Quantitative comparison of FinerCut similarity metrics, including cosine similarity and MSE, evaluated using FID, CLIP, DreamSim, and GenEval. The results confirm that MSE outperforms cosine similarity, which is consistent with the finding of ours.($\uparrow / \downarrow$ denotes that a higher / lower metric is favorable).}
    \label{append_tab:finercut_cosine}
    \centering
    \resizebox{\textwidth}{!}{
    \begin{tabular}{cccccccccccc}
        \toprule
        \multirow{2}{*}{Method} & Sparsity & \multirow{2}{*}{FID $\downarrow$} & \multirow{2}{*}{CLIP $\uparrow$} & \multirow{2}{*}{DreamSim $\uparrow$} & \multicolumn{7}{c}{GenEval $\uparrow$}    \\
        \cmidrule(lr){6-12}
        & (\%) & & & & Single & Two & Count. & Colors & Pos. & Color attr. & Overall \\
        \midrule
         Dense & 0.0 & 22.89 & 0.314 & 1.0 & 0.988 & 0.616 & 0.475 & 0.795 & 0.108 & 0.255 & 0.539 \\
        \cmidrule(lr){1-12}
        \multirow{3}{*}{FinerCut (cos.)} & 26.3 & 19.64 & 0.311 & 0.745 & 0.925 & 0.394 & 0.363 & 0.657 & 0.060 & 0.085 & 0.414 \\
                                         & 32.2 & 21.16 & 0.306 & 0.689 & 0.875 & 0.346 & 0.322 & 0.620 & 0.043 & 0.058 & 0.377 \\
                                         & 41.7 & 21.84 & 0.306 & 0.671 & 0.847 & 0.275 & 0.256 & 0.625 & 0.053 & 0.033 & 0.348\\
        \midrule
        \multirow{3}{*}{FinerCut (MSE)} & 26.3 & 20.15 & 0.315 & 0.800 & 0.947 & 0.465 & 0.394 & 0.737 & 0.103 & 0.105 & 0.458 \\
                                          & 32.2 & 20.19 & 0.313 & 0.775 & 0.903 & 0.409 & 0.344 & 0.697 & 0.078 & 0.128 & 0.426 \\
                                          & 41.7 & 19.93 & 0.312 & 0.741 & 0.841 & 0.306 & 0.306 & 0.628 & 0.050 & 0.073 & 0.367\\
        \midrule
        \multirow{3}{*}{\textbf{Skrr (Ours)}} & 27.0 & 20.15 & 0.315 & 0.800 & 0.956 & 0.434 & 0.425 & 0.763 & 0.095 & 0.145 & 0.471 \\
                                               & 32.4 & 20.19 & 0.313 & 0.775 & 0.928 & 0.397 & 0.413 & 0.774 & 0.100 & 0.118 & 0.455 \\
                                               & 41.9 & 19.93 & 0.312 & 0.741 & 0.913 & 0.410 & 0.450 & 0.755 & 0.055 & 0.068 & 0.442\\
        \bottomrule
        \end{tabular}
        }
\end{table*}

\subsection{Additional qualitative results}
\begin{figure}[p]
  \centering
\includegraphics[width=0.95\linewidth]{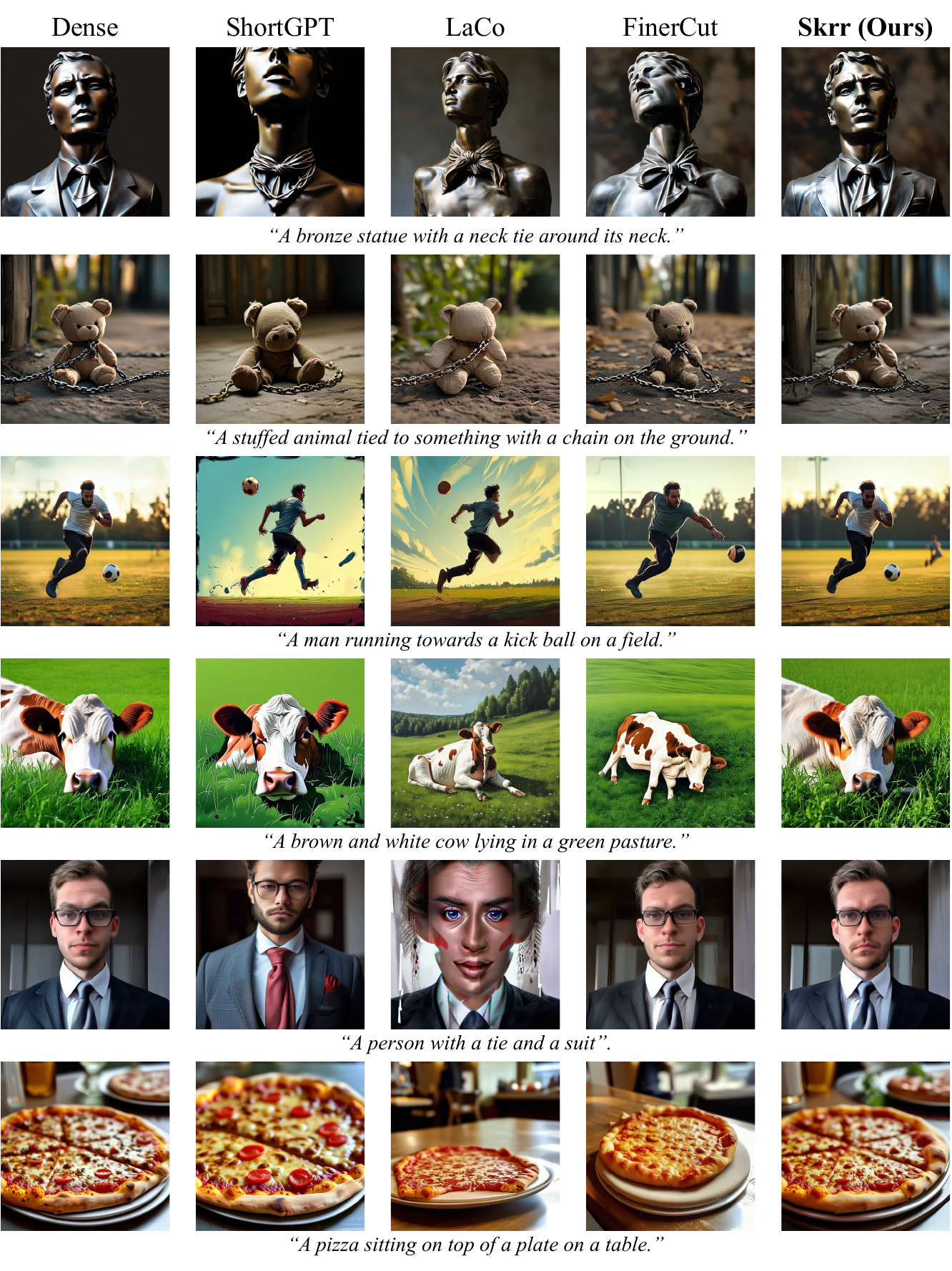}
\vspace{-1em}
\caption{
Qualitative comparison of images generated by models compressed to sparsity range 20\%-30\% (sparsity level 1) using ShortGPT, LaCo, FinerCut, and Skrr, alongside images generated by dense models. Skrr demonstrates a remarkable ability to produce images closely resembling those from the dense model in the majority of cases.}
\label{append_fig:quali_1}
\end{figure}

\begin{figure}[p]
  \centering
\includegraphics[width=0.95\linewidth]{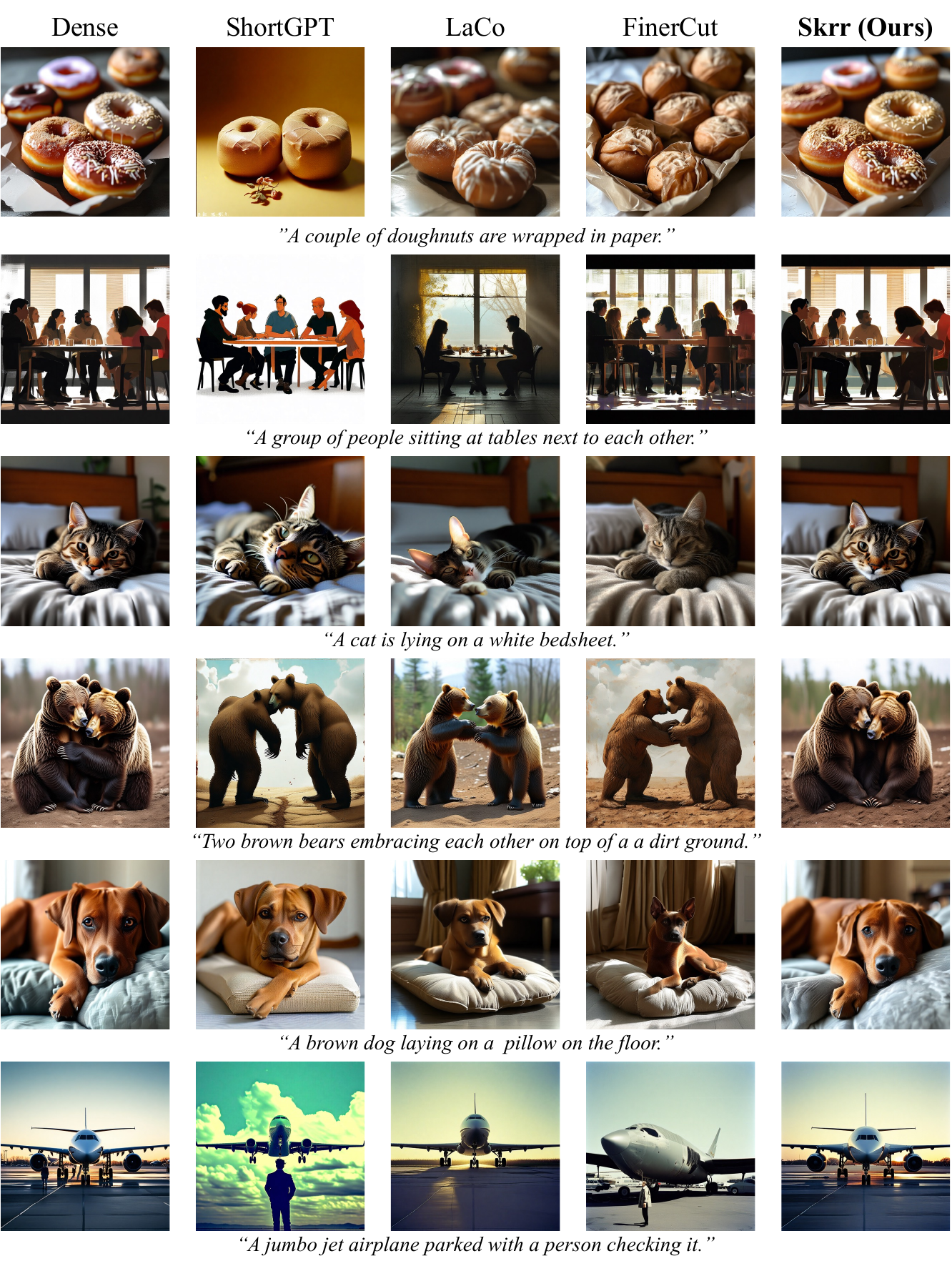}
\vspace{-1em}
\caption{
Qualitative comparison of images generated by models compressed to sparsity range 30\%-40\% (sparsity level 2) using ShortGPT, LaCo, FinerCut, and Skrr, alongside images generated by dense models. Skrr demonstrates a remarkable ability to produce images closely resembling those from the dense model in the majority of cases.}
\label{append_fig:quali_2_1}
\end{figure}

\begin{figure}[p]
  \centering
\includegraphics[width=0.95\linewidth]{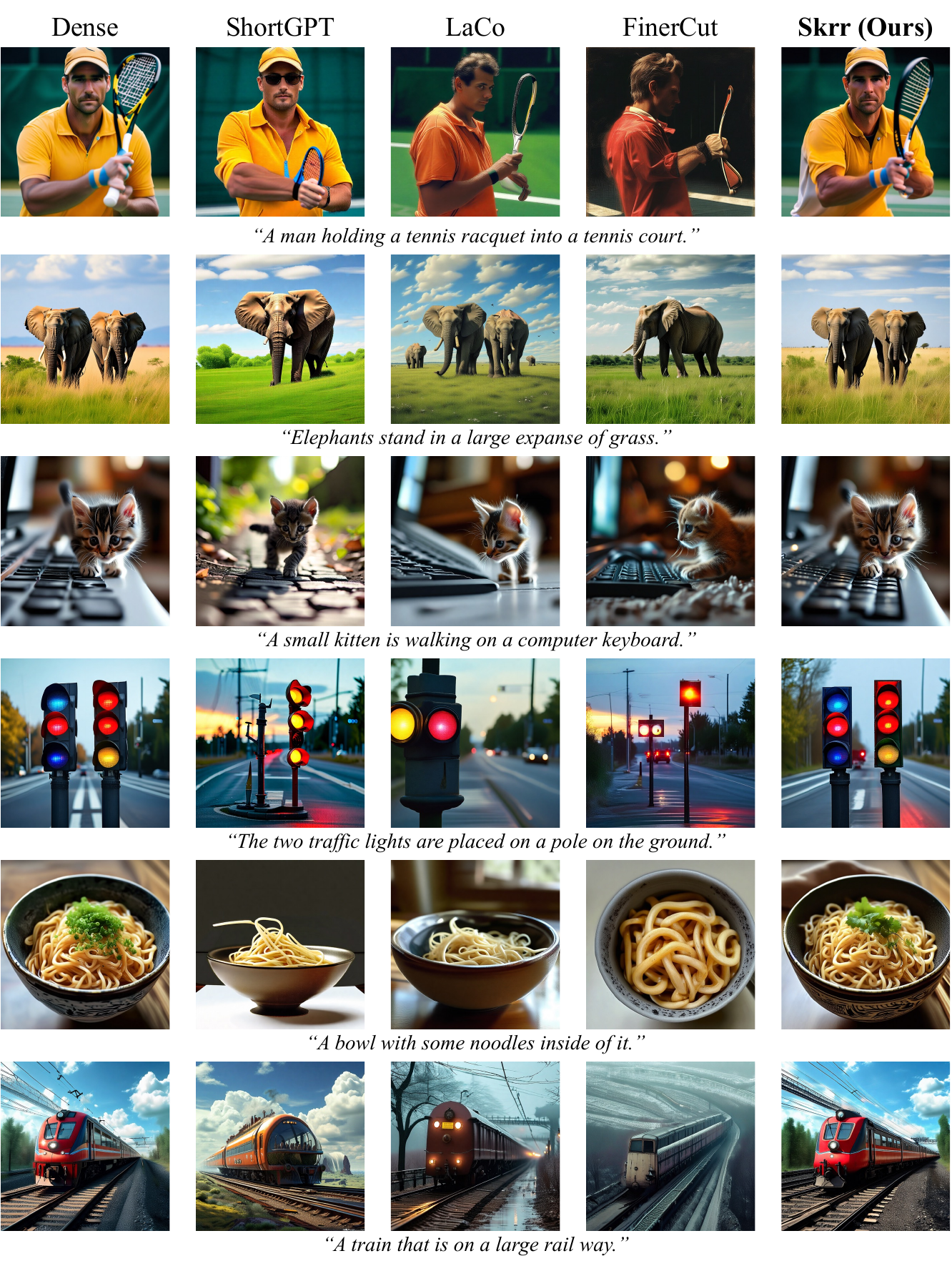}
\vspace{-1em}
\caption{
Qualitative comparison of images generated by models compressed to sparsity range 30\%-40\% (sparsity level 2) using ShortGPT, LaCo, FinerCut, and Skrr, alongside images generated by dense models. Skrr demonstrates a remarkable ability to produce images closely resembling those from the dense model in the majority of cases.}
\label{append_fig:quali_2_2}
\end{figure}

\begin{figure}[p]
  \centering
\includegraphics[width=0.95\linewidth]{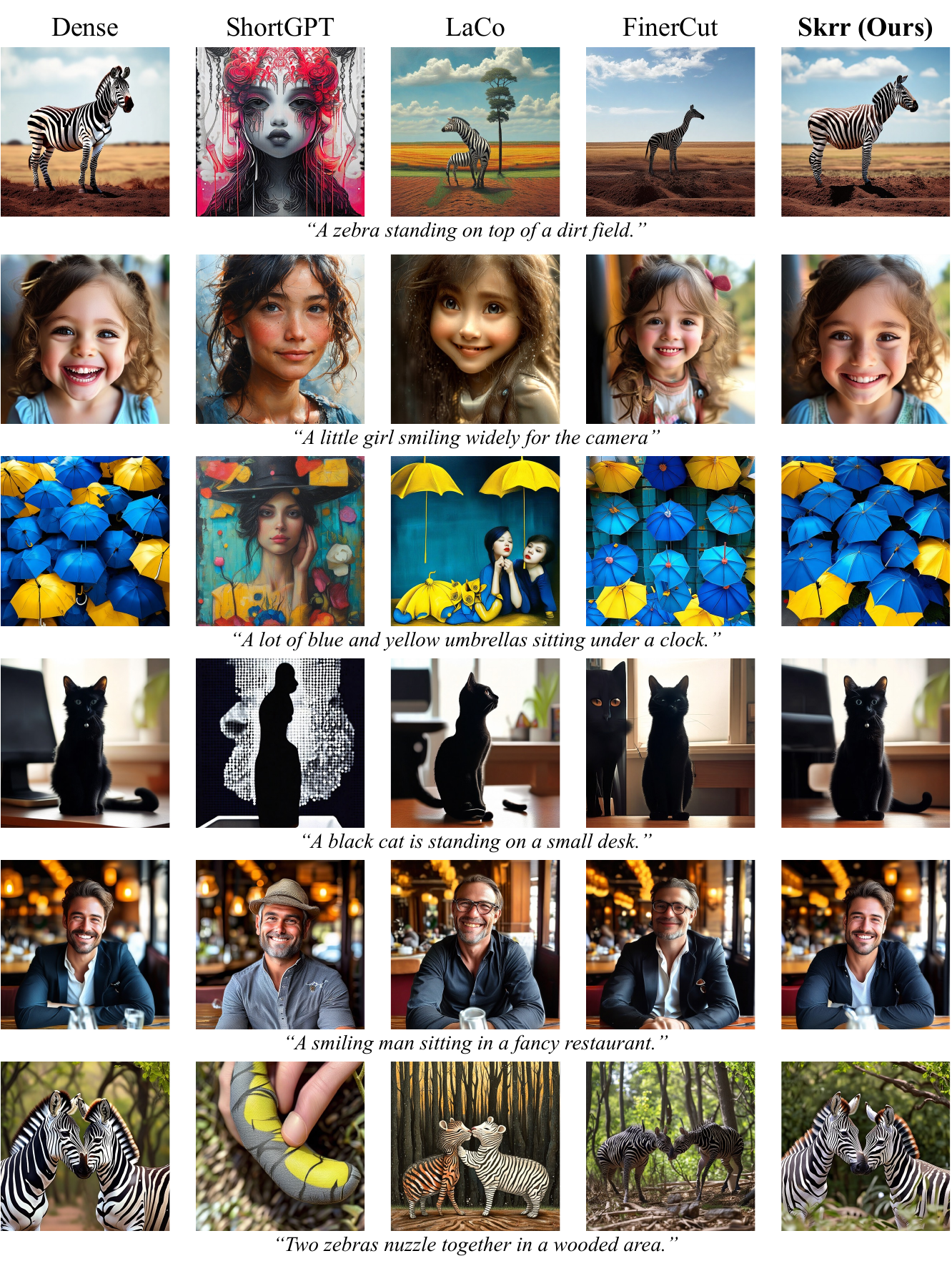}
\vspace{-1em}
\caption{
Qualitative comparison of images generated by models compressed to sparsity range 40\%-50\% (sparsity level 3) using ShortGPT, LaCo, FinerCut, and Skrr, alongside images generated by dense models. Skrr demonstrates a remarkable ability to produce images closely resembling those from the dense model in the majority of cases.}
\label{append_fig:quali_3_1}
\end{figure}

\begin{figure}[p]
  \centering
\includegraphics[width=0.95\linewidth]{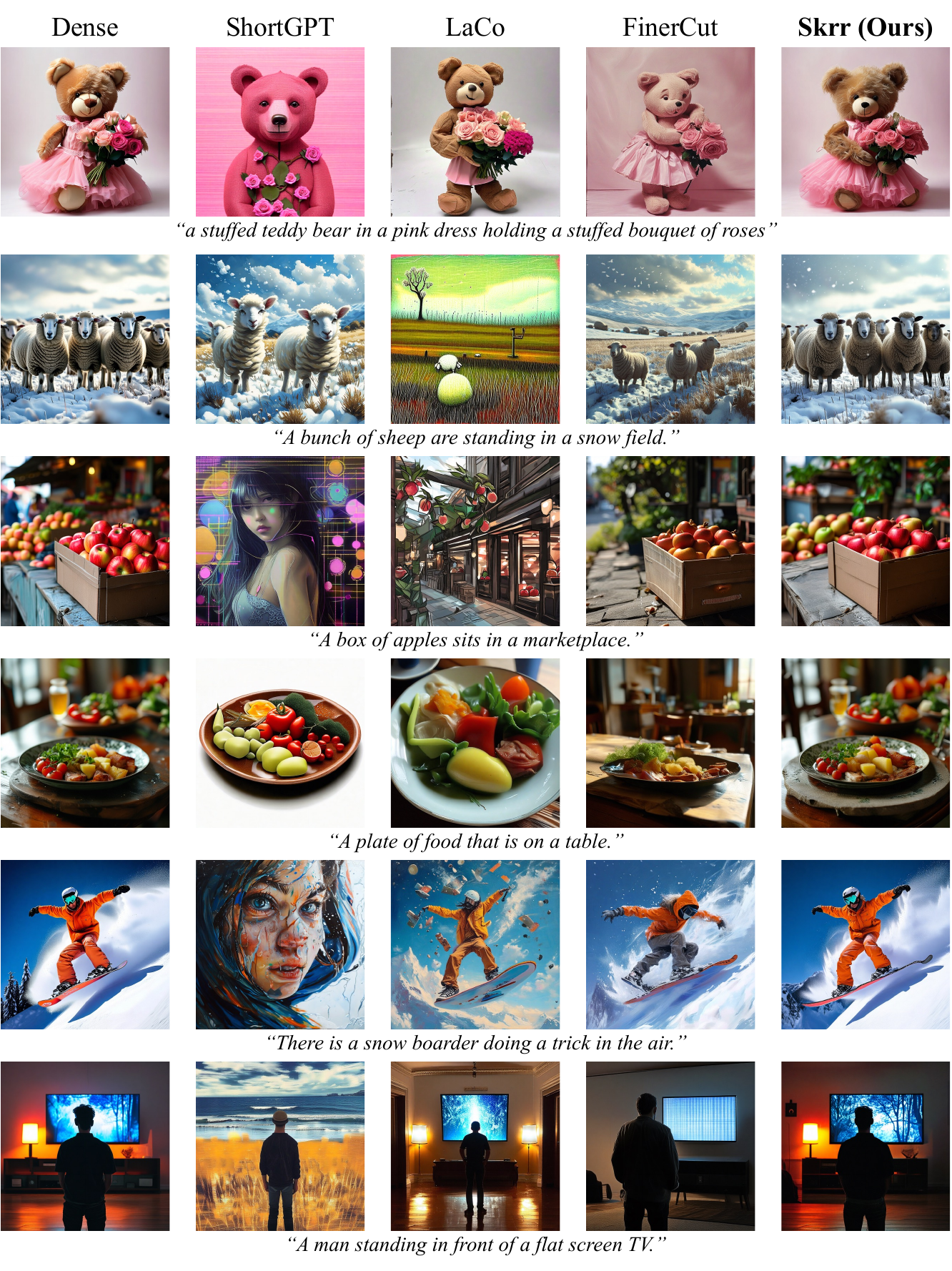}
\vspace{-1em}
\caption{
Qualitative comparison of images generated by models compressed to sparsity range 40\%-50\% (sparsity level 3) using ShortGPT, LaCo, FinerCut, and Skrr, alongside images generated by dense models. Skrr demonstrates a remarkable ability to produce images closely resembling those from the dense model in the majority of cases.}
\label{append_fig:quali_3_2}
\end{figure}

In this section, we present additional qualitative results to further illustrate the performance of our approach. We define the model compressed with 20\%-30\% sparsity as sparsity level 1, 30\%-40\% as sparsity level 2, and 40\%-50\% as sparsity level 3. The qualitative comparison for sparsity level 1 is shown in Fig.~\ref{append_fig:quali_1}, while comparisons for sparsity level 2 are presented in Fig.~\ref{append_fig:quali_2_1} and Fig.~\ref{append_fig:quali_2_2}. Finally, Fig.~\ref{append_fig:quali_3_1} and Fig.~\ref{append_fig:quali_3_2} illustrate the comparisons for sparsity level 3. The results demonstrate that Skrr generates outputs that are more closely aligned with the prompts and more consistent with the outputs of the dense model compared to other baselines.
All presented results were generated using PixArt-$\Sigma$ and are part of a dataset comprising 30,000 images, which were produced for FID measurements.

\begin{figure}[p]
  \centering
\includegraphics[width=0.95\linewidth]{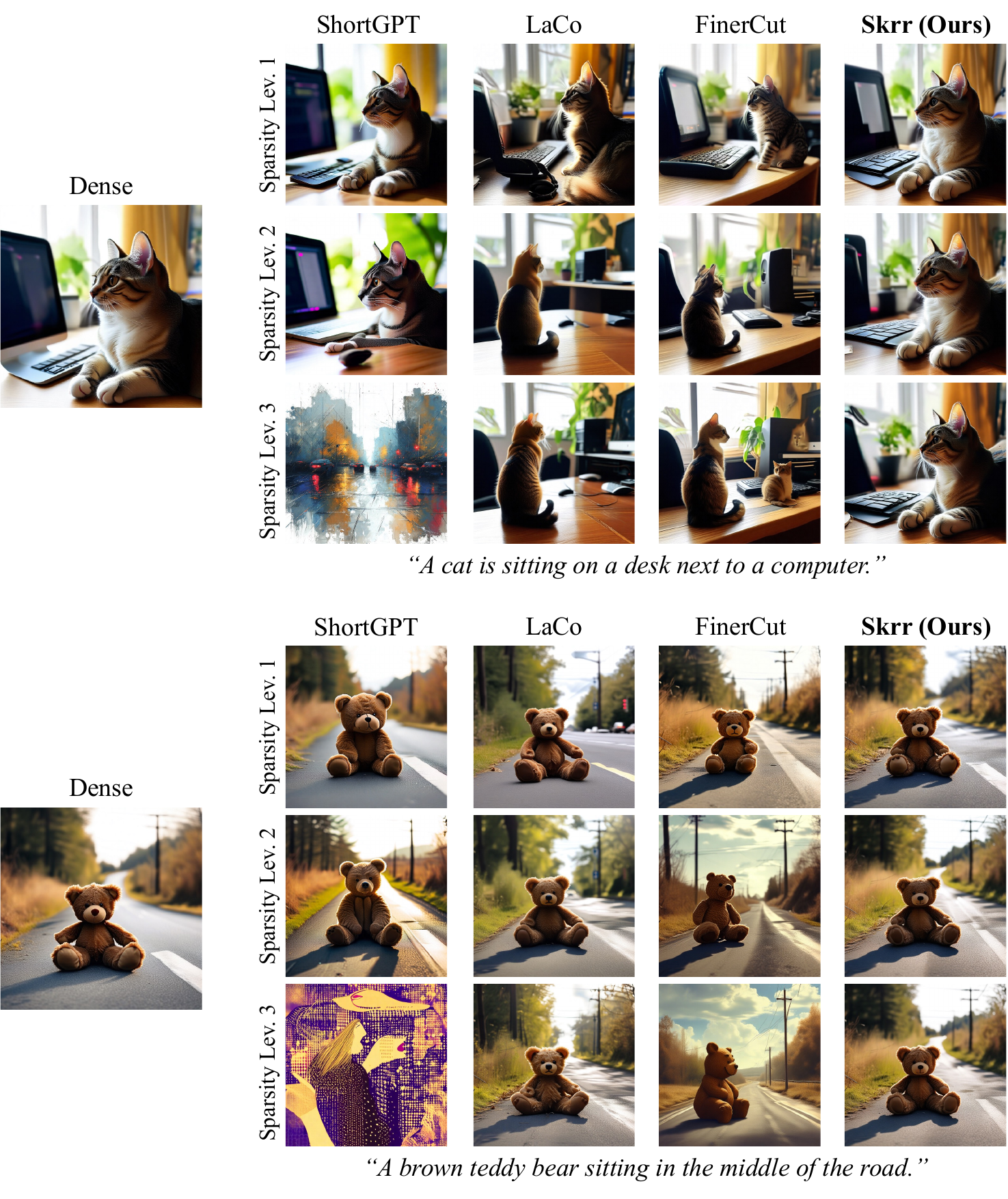}
\vspace{-1em}
\caption{
Qualitative comparison of images generated by models compressed to various sparsity level using ShortGPT, LaCo, FinerCut, and Skrr, alongside images generated by dense models. Skrr demonstrates a remarkable ability to produce images closely resembling those from the dense model in all the sparsity levels. Notably, the lighting and composition of objects in the images retains even after harsh pruning to the text encoder.}
\label{append_fig:quali_sparse_lev_1}
\end{figure}

\begin{figure}[p]
  \centering
\includegraphics[width=0.95\linewidth]{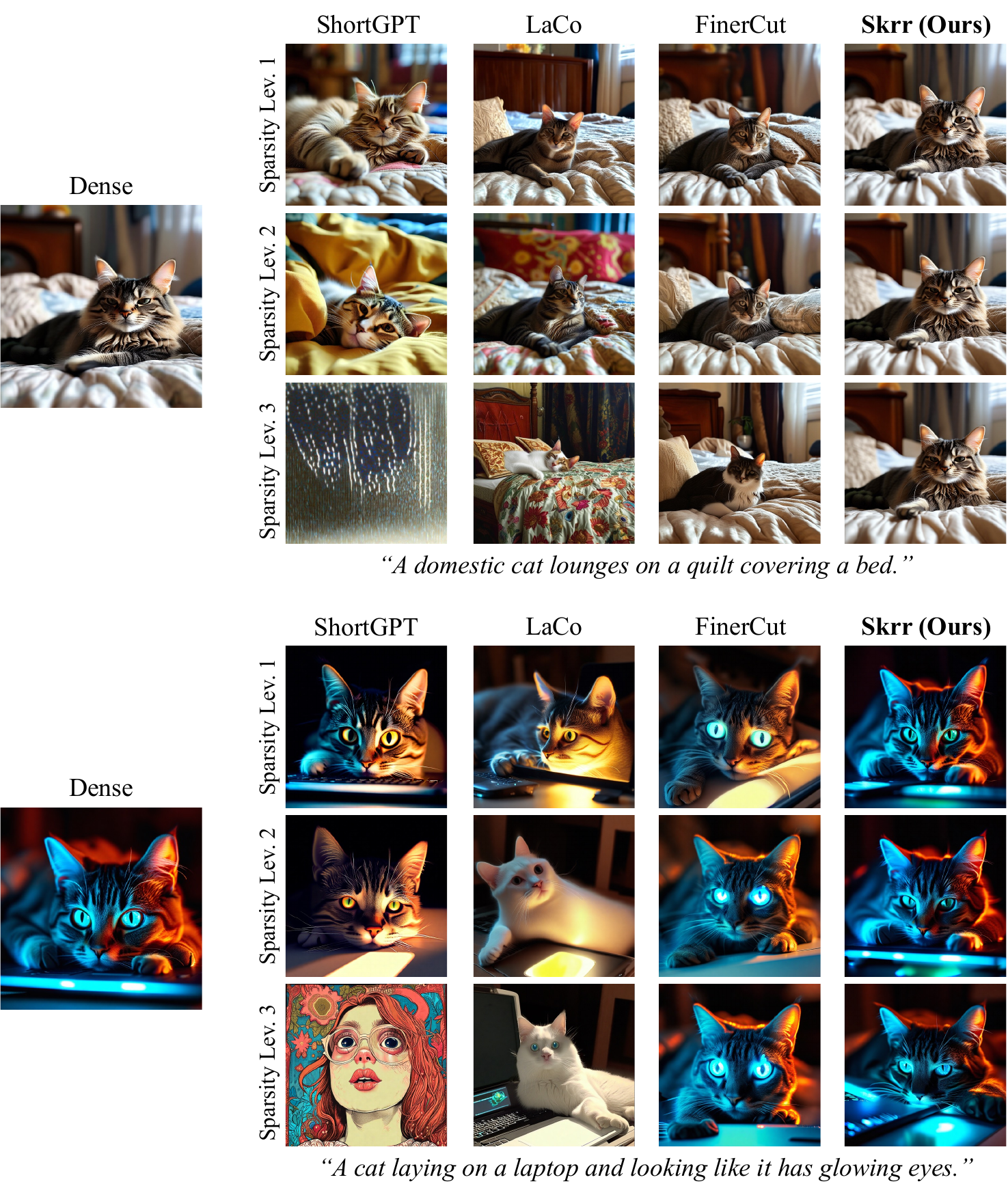}
\vspace{-1em}
\caption{
Qualitative comparison of images generated by models compressed to various sparsity level using ShortGPT, LaCo, FinerCut, and Skrr, alongside images generated by dense models. Skrr demonstrates a remarkable ability to produce images closely resembling those from the dense model in all the sparsity levels. Notably, the lighting and composition of objects in the images retains even after harsh pruning to the text encoder.}
\label{append_fig:quali_sparse_lev_2}
\end{figure}

Additionally, we provide a qualitative comparison of the text encoders compressed by each compression method across all sparsity levels, juxtaposed with the images generated by the dense model. By comparing images generated with the same seed at sparsity levels 1, 2, and 3, as defined above, we can assess the degree of deviation from the original image as the sparsity increases. These comparisons are illustrated in Fig.~\ref{append_fig:quali_sparse_lev_1} and Fig.~\ref{append_fig:quali_sparse_lev_2}. The results reveal that, while the baseline methods occasionally generate images similar to those from the dense model at low sparsity, the differences become more pronounced as sparsity levels increase. In contrast, the model compressed using Skrr consistently maintains a high degree of similarity to the original images across all sparsity levels, demonstrating its robustness.

\section{Limitations}
In this section, we address the limitations of Skrr. While Skrr effectively preserves Text-to-Image (T2I) performance during pruning, its performance deteriorates noticeably at extreme sparsity levels ($> 50\%$). This degradation is likely due to a significant reduction in the representational capacity of the text encoder as the number of parameters becomes excessively limited. However, such a sparsity could still provide additional memory efficiency through complementary techniques such as weight quantization. Another limitation is that Skrr does not achieve performance improvements beyond that of the original dense model. While pruning the text encoder can enhance certain image quality metrics, such as improving FID scores, we observed a consistent decline in performance on benchmarks like CLIP score and GenEval as sparsity increased. Addressing this performance drop and ensuring robustness across benchmarks would be an interesting future work.

\end{document}